\documentclass[11pt]{article}
\usepackage[letterpaper,margin=0.875in]{geometry}
\usepackage{lmodern}
\usepackage{array}
\usepackage[T1]{fontenc}
\usepackage{natbib}
\usepackage{graphicx}
\usepackage{booktabs}
\usepackage{amsmath,amssymb}
\usepackage{multirow}
\usepackage{xcolor}
\usepackage{url}
\usepackage[hidelinks]{hyperref}
\usepackage[expansion=false]{microtype}
\usepackage{float}
\newcommand{\bench}{\textsc{TypedBench}}%
\newcommand{\jev}{\textsc{Jev}}
\newcommand{\laya}{\textsc{Laya}}
\newcommand{\kev}{\textsc{Kev}}

\begin{document}

\title{\textsc{TypedBench}: A Benchmark for Calibration, Framing Sensitivity, and Cost in System One Decision Models}
\author{{\small\textbf{Rahul Sharma}$^{1, 2}$ \quad \textbf{Andrew B. Duncan}$^{3}$ \quad \textbf{Gaétan Marceau Caron}$^{4}$ \quad \textbf{Sebastian J. Vollmer}$^{1, 2}$}\\[0.8em]
{\small $^{1}$DFKI GmbH \quad $^{2}$RPTU Kaiserslautern \quad $^{3}$Imperial College London \quad $^{4}$Mila - Quebec AI Institute}}
\date{}
\maketitle

\begin{abstract}
System One models output calibrated probabilities over typed answers such as categorical choices, ordinal levels, or binary outcomes, via a non-generative interface. Software can act on these probabilities through thresholds, cost-weighted choices, and escalation rules. Consequently, if these probabilities are miscalibrated or wording-sensitive, the software may take unintended actions leaving human operators with no textual rationale to inspect. Current evaluations largely report accuracy and calibration on public classification datasets without a clear reference. We present $\texttt{TypedBench}$, a benchmark for typed decision models like Jev built from seven policy-labelled generators and nine evaluation suites. We report accuracy as median and range across paraphrases, and calibration error relative to the finite-sample noise floor of a matched, perfectly calibrated predictor. We assess probability quality through selective prediction, ordinal proper scoring rules, and realised cost under asymmetric cost matrices. We evaluate a hosted model, an open encoder, and a family of open decoders spanning 0.8B–9B parameters on identical items. The hosted model follows the stated policy but is wording-sensitive and systematically underconfident; under asymmetric costs, using its probabilities can be worse than taking its top answer. Decoders route exactly and are unaffected by co-asked questions, but apply policies only partially and are slow as options or questions are added. The encoder is least accurate on policy questions and degrades with more options. Overall, typed decision models must be evaluated jointly on policy adherence, wording robustness, probability quality, and induced decision outcomes.
\end{abstract}

\section{Introduction}

A new class of non-generative AI interfaces, termed "System One" models, outputs direct probability distributions over candidate answers rather than generating free-form text. Given a \emph{state} and a set of typed \emph{questions}, they return a probability distribution over user-defined answers: a \emph{choice} among up to 255 named options, a \emph{score} on an ordered rubric of 2--10 levels, or a \emph{noul} (a yes/no) probability. All questions are answered in a single forward pass, and the output requires no parsing. Such models are being called \emph{System One models}, after the fast, automatic mode of dual-process psychology \citep{kahneman2011,typesafe2026jev}. \jev{}, a closed-source model from Typesafe AI introduced this interface. Simultaneously, an open-weight encoder \citep{convai2026laya} and an open-weight decoder family \citep{palmer2026kev} adopted the same interface. The appeal is speed, input-only billing, and most importantly \emph{calibrated} probabilities that downstream systems can threshold, price, or route on \citep{typesafe2026jev}.

The claim that these System One models are calibrated is important for deployment but existing evaluations do not directly test it. The hosted model's evaluation compares against the average of two frontier language models rather than against ground truth; the open encoder's reported results use a checkpoint fine-tuned on the benchmark's own training split; and independent tests reported a 33-point accuracy swing from rewording a question \citep{xenospectrum2026}, an expected calibration error roughly four times its finite-sample noise floor \citep{scienthoon2026calib}, and a severe drop in encoder performance when evaluating more than twenty options \citep{mehra2026jevbench}. Finally, most public results are reported on datasets (e.g.\ Banking77, CLINC150, SST, AG~News) that are widely represented in pretraining corpora \citep{casanueva2020banking77,larson2019clinc}.

Benchmarks for non-generative decision models differ from language-model benchmarks; there is no answer string to match, so evaluation essentially reduces to three questions. \emph{Are the probabilities meaningful?} \emph{Is the accuracy stable under rewording?} \emph{Are the test items free of training-set contamination?} Based on these questions, \bench{} contributes the following:

\begin{enumerate}\setlength\itemsep{0pt}
\item \textbf{Policy-labelled, contamination-resistant construction with derivable labels.} We introduce seven  programmatic generators that produce fresh items whose labels are derived from explicit rules. Paired controls test unknowable cases, label noise, distractors, none-correct cases, and surface perturbations. 

\item \textbf{An evaluation protocol for typed probabilities.}
We report accuracy across friendly and adversarial framings, calibration relative to a matched finite-sample null ECE, selective prediction, ordinal scores, and decision value under asymmetric costs.

\item \textbf{A cross-architecture evaluation of six checkpoints.} We evaluate a hosted model, two encoder checkpoints, and three decoder sizes on identical items. The resulting failure modes are complementary: the hosted model is sensitive to wording and systematically underconfident; the encoder degrades with structural complexity and option count; and the decoders are comparatively stable but incur substantial latency and sometimes only partially apply the stated policy.
\end{enumerate}
The resulting benchmark is intended as a measurement layer for typed decision interfaces. Rather than producing a leaderboard score, it exposes the dimensions on which a model can succeed or fail when its probabilities are consumed by software. 

\section{Related Work}

\paragraph{Zero-shot classification.} BTZSC \citep{aarab2026btzsc} compares 38 checkpoints on 22 public datasets, finding rerankers ahead of instruction-tuned LLMs. \citet{rodrigues2026intent} show that fine-tuned RoBERTa outperforms zero-shot LLMs by 11.8 points on ATIS while tying on CLINC150, with the LLM leading only on out-of-scope recall. Additionally, \citet{lee2026finetuning} show that label wording substantially alters zero-shot behaviour. All three benchmarks reuse public datasets and report accuracy without evaluating probabilities; \bench{} differs in both respects.

\paragraph{Calibration and uncertainty.} KalshiBench \citep{nel2025kalshibench} evaluates models on prediction market outcomes that occurred after their training cutoffs, finding systematic overconfidence. \citet{tao2025uncertainty} evaluate 80 models and show that good calibration does not imply good error ranking (i.e., ordering predictions so that incorrect outputs receive the lowest confidence). Similarly, \citet{deoliveira2025calibration} show that off-the-shelf generative models exhibit high calibration error (24--47\% ECE), though these can be substantially reduced via post hoc scaling. However, these works score verbalised or token-level confidence from generative models and do not evaluate a model that natively returns a distribution over typed options, and do not report a finite-sample noise floor for ECE \citep{naeini2015ece,guo2017calibration,kumar2019verified}.

\paragraph{Prompt and framing sensitivity.} PromptSET \citep{razavi2025promptset} established wording as a major variance factor. \citet{hua2025flaw} show that much of the reported sensitivity in generative models is an artifact of the matching of free-text responses. As typed models have no free-text answer to mis-match, they provide a cleaner setting for measuring prompt sensitivity. Building on this, \citet{cox2025mapping} decompose uncertainty into a wording-attributable share, which we adopt as a reporting target.

\paragraph{Ordinal judging, guardrails, out-of-scope, order.} \citet{li2025scoringbias} identify rubric-order and score-ID bias in absolute-score judging, while \citet{rao2026autorubric} introduce weighted kappa for ordinal evaluation. Regarding safety controls, \citet{fomin2026lodo} shows that guardrail classifiers overstate generalisation by 8--16 AUC points under leave-one-dataset-out evaluation, and \citet{goehausen2026gate} argue for evaluation at a single global operating point. For out-of-scope (OOS) detection, \citet{zaera2025oos} show that detection quality depends on whether an explicit OOS class is provided, while \citet{yin2025fragile} document quality-dependent order effects. \bench{} operationalises each of these failure modes through  paired control arms evaluated on identical items.

\paragraph{Contamination and cascades.} Surveys of contamination-free evaluation \citep{cheng2025contamination,chen2025dynamic} catalog data updating, rewriting, and prevention strategies for generative benchmarks; however, none treat policy-dependent labels. In cost-aware routing, \citet{kotte2026ucci} shows that calibrated cascade routing cuts cost by 31\% at a fixed F1 score; our benchmark evaluates this same cost-saving trade-off offline via a hybrid threshold sweep. Finally, \citet{choi2026conformal} study calibration under label shift; our prior-shift arm (evaluating under changing class base rates) is, to our knowledge, the first such measurement for decision models.

\section{Benchmark Design}\label{sec:design}

\subsection{Decision Model Interface}
A model satisfies the decision model interface by implementing a function $(s,q) \mapsto p$: given an input context state $s$ and a question $q$ belonging to a task primitive $\{\text{choice},\text{score},\text{noul}\}$ with options $\mathcal{K}_q$, it returns a probability vector $p\in\Delta({\mathcal{K}_q)}$ in one call and without generating text. To ensure valid execution, model adapters publish their operational \emph{capabilities} such as supported primitives, maximum option count, state token limit, abstention, and batching.

\subsection{Benchmark Tasks as Policy-Driven Programmatic Generators}\label{sec:data}
\paragraph{Generators.} Rather than relying on public datasets that are vulnerable to test-set contamination, our benchmark framework evaluates decision models using a suite of programmatic task generators with policy-driven ground truth. The framework spans seven operational tasks given in \autoref{tab:generators}. More details in \autoref{app:gen}.

\begin{table}[t]
\centering\footnotesize
\caption{The seven generators and the capability each isolates.}
\label{tab:generators}
\setlength{\tabcolsep}{3pt}
\begin{tabular}{@{}>{\raggedright\arraybackslash}p{1.9cm}>{\raggedright\arraybackslash}p{2.8cm}>{\raggedright\arraybackslash}p{2.85cm}@{}}
Generator & Questions & Capability isolated \\
\midrule
support tickets & queue (up to 12), anger, priority (5 levels) & applying a three-term policy \\
phishing email & phishing, attack class (5-way), urgency (4 levels), five sub-questions & decomposing a judgement into evidence \\
RAG relevance & best passage among $K$ ($K\le255$), relevant, relevance (4 levels) & choosing among many options with hard distractors \\
log triage & severity (4 levels), owning team (5), page on-call & a lookup table and a two-field rule in a JSON record \\
policy compliance & compliant, violated clause or none, severity (3 levels) & matching a request to numbered clauses \\
guardrail intent & intent (4), block, risk (4 levels) & separating attack mechanism from harm \\
multilingual tickets & as support tickets, in de/es/fr/hi & the same policy across languages \\
\bottomrule
\end{tabular}
\end{table}

\paragraph{Control Streams and Perturbations.} To evaluate model robustness under realistic noise, pseudo-randomness is partitioned into independent content and control streams. This guarantees that each stress-test variation pairs item-for-item against its corresponding clean baseline item. The framework applies five distinct control-arm perturbations: \emph{unknowable} (removing tier context so two priority levels become equally valid), \emph{label noise} (corrupting 5\% of labels), \emph{distractors} (injecting noise at 0.25 or 0.5 density), \emph{none-correct} (withholding the ground-truth option), and \emph{surface perturbations} (introducing casing irregularities, whitespace shifts, 2\%--5\% typos, and Cyrillic homoglyphs). To guarantee reproducibility, datasets are hash-verified and generator code is version-controlled.

\paragraph{Label Derivability.} Ground truth labels are generated programmatically rather than annotated by hand. Each task template explicitly tags underlying facts using rule defined vocabulary, including whether a ticket involves money at stake or an operational blocker, or whether a guardrail prompt has a harmful objective. The ground truth label is then derived deterministically by executing the policy rules over these declared facts. To ensure data quality, a two stage audit is applied prior to release. First, automated unit tests verify that labels match the underlying policy logic and confirm that generated texts contain the necessary cues. Second, a frontier LLM is evaluated on the state and prompt text to validate that the questions are unambiguous and solvable (\autoref{app:audits}).

\subsection{Evaluating Model Robustness Across Framing Variants}\label{sec:framing}
To evaluate model robustness against prompt variations, each item is evaluated under multiple framing conditions while holding the ground-truth label constant. Prompts include canonical wording, five human-written paraphrases, three criteria variants (label-only, with negative definitions, or deliberately vague), a corruption control with swapped option descriptions, and adversarial wordings written to mislead, all fixed before evaluation to reduce accuracy. We define non-adversarial variants (canonical, paraphrased, and explicit criteria) as \emph{friendly framings}. Across these variants, we report accuracy as the median over friendly framings alongside $[\min, \max]$ bounds and show every adversarial wording separately. To measure prediction stability, we compute the mean pairwise Jensen--Shannon divergence of the returned output distributions and track the argmax flip rate across framings. Additionally, to test for positional bias, option orders are permuted three times per item, and a $\chi^2$ test is applied to assess index selection distributions.

\subsection{Metrics}\label{sec:metrics}
\paragraph{Calibration.} Top-label ECE with 15 equal-width bins, the binning-free smooth calibration error, Brier score with its Murphy decomposition \citep{brier1950,murphy1973brier}, and NLL with probabilities floored at $10^{-4}$. Every ECE is compared with its \emph{null ECE}, the expected ECE of a perfectly calibrated model with the same confiedence histogram, from 200 label resamplings $y_i\sim\mathrm{Bernoulli}(c_i)$; we report the ECE ratio ECE/null as a descriptive measure of excess calibration error. A temperature $T$ is fitted by NLL on one half of the items and evaluated on the other; $T>1$ is over-confidence, and the fit is declared degenerate when the fit half is fully correct.

Every ECE is compared with its \emph{null ECE}, the expected ECE of a perfectly calibrated model with the same confidences, from label resamplings $y_i\sim\mathrm{Bernoulli}(c_i)$; we report the ECE ratio ECE/null as a descriptive measure of excess calibration error (1 on average under perfect calibration, up to about 1.5 by chance).

\paragraph{Selective prediction.} Risk--coverage curves, AURC and coverage at $\le\!5\%$ risk \citep{geifman2017selective}.
\paragraph{Ordinal.} Exact and off-by-one accuracy, MAE of the top answer and of the expected level, quadratic weighted kappa \citep{cohen1968kappa}, the ranked probability score \citep{epstein1969rps,gneiting2007proper}, monotonicity of the expected level along controlled severity ladders, and Spearman agreement across 3-, 5- and 10-level rubrics.
\paragraph{Decision value.} With cost matrix $C$ (true $\times$ predicted), realised cost per $10^4$ decisions under the argmax and Bayes ($\arg\min_j \sum_t p_t C_{tj}$) policies; the \emph{value of calibration} is cost(argmax) $-$ cost(Bayes).
\paragraph{Efficiency.} Client-observed latency for hosted models with the route recorded and latency on stated hardware for local models.

\subsection{Suites}
Table~\ref{tab:suites} lists the nine suites designed to test typed decision models. Each finding in \S\ref{sec:results} names the suites that support it. 

\begin{table}[t]
\centering\footnotesize
\caption{The nine evaluation suites: the question each answers, what it varies, and its main metric.}
\label{tab:suites}
\setlength{\tabcolsep}{3pt}
\renewcommand{\arraystretch}{0.95}
\begin{tabular}{@{}l >{\raggedright\arraybackslash}p{2.2cm} >{\raggedright\arraybackslash}p{2.65cm} >{\raggedright\arraybackslash}p{1.75cm}@{}}
\toprule
 & Asks & Varies & Metric \\
\midrule
A & Do confidences match accuracy? & unknowable and label-noise arms & ECE/null, $T$, AURC \\
B & Is accuracy stable under rewording? & paraphrases, criteria variants, adversarial wordings, option order & accuracy range, JSD, flips \\
C & Does it scale? & options $K=2$--255; state length 128--30k tokens & accuracy, latency \\
D & Is it robust? & distractors, perturbations, none-correct, prior shift & accuracy change, abstention, AUROC \\
E & Do co-asked questions interfere? & 2--20 co-questions & answer changes, latency \\
F & Are scores ordered? & severity ladders; 3/5/10-level rubrics & monotonicity, Spearman \\
G & Are yes/no answers coherent? & negation; choice vs.\ yes/no; another task's threshold & complement sum, JSD \\
H & Can confidence route? & escalation threshold to a second model & accuracy vs.\ cost \\
I & Does acting on probabilities pay? & cost matrix; under- : over-prioritising 1--20 & cost per $10^4$ \\
\bottomrule
\end{tabular}
\end{table}

\subsection{Statistics and audits}
The sources of variance are the item sample, option order, wording and version drift, all driven by seed variability. Differences are tested by item-level paired bootstrap (10{,}000 resamples) and Holm-corrected McNemar's test \citep{holm1979} (\autoref{app:stats}). At $n=500$ the 95\% interval on an accuracy near 0.5 is about $\pm4.4$ points. Audits run on every manifest: state-only and options-only short circuits (flagged when either exceeds the prior by 10 points), leakage of teh true option's name or description into the state, the positional-bias test of \S\ref{sec:framing}, and a label-noise control in which the accuracy against the corrupted labels must stay near zero and confidence must match clean items, since only the label changes. A 200-item canary is run at the start of each hosted session and compared to the last accepted run; results across a drift alarm are never merged.

\begin{figure*}[t]
\centering
\includegraphics{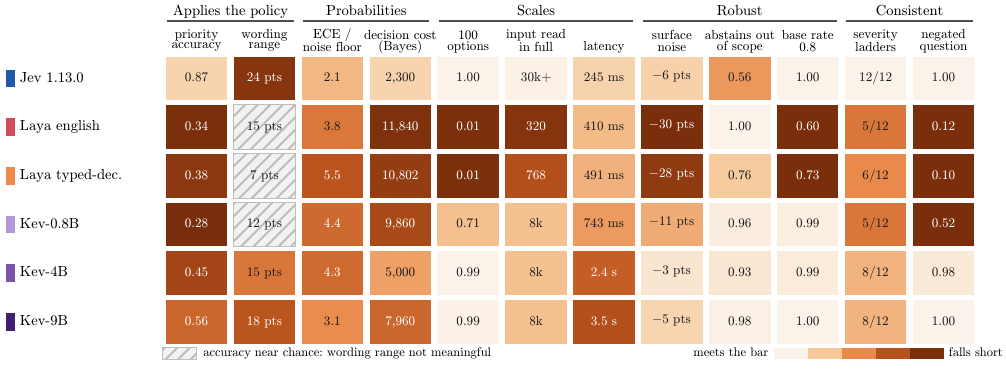}
\caption{Where each model falls short. Rows are checkpoints, columns the questions the benchmark asks. Each cell gives the measured value, shaded from light (meets the bar) to dark (falls short) on a fixed scale per column, so the shading does not depend on which models are compared. \emph{Applies the policy}: median accuracy on ticket priority over rewordings of the rule, and its range across them (hatched where accuracy is near chance, so the range says little). \emph{Probabilities}: calibration error as a multiple of its finite-sample null ECE (median over priority, urgency, phishing and attack class), and the cost per $10^4$ priority decisions made from the probabilities when under-prioritising costs twice as much as over-prioritising. \emph{Scales}: accuracy with 100 options, the longest input read without truncation, and median latency with 21 questions per request. \emph{Robust}: largest accuracy drop under surface noise, abstention on out-of-scope items, and accuracy when the base rate of anger shifts to 0.8. \emph{Consistent}: severity ladders kept in order, and accuracy on a reworded negation of the anger question. Accuracy columns are scaled from chance or the majority answer to 1.}
\label{fig:story}
\end{figure*}

\section{Experiments and Results}\label{sec:experiments}
\subsection{Evaluation Setup}\label{sec:setup}
\paragraph{Models.} \jev{} 1.13.0 \citep{typesafe2026jev}, hosted, reached through the first-party API; \laya{} 0.3.4 \citep{convai2026laya}, a 421M ModernBERT-large \citep{warner2024modernbert} encoder with a decision head, in its English and typed-decisions checkpoints, run locally in bf16; and \kev{} \citep{palmer2026kev}, a LoRA-plus-pointer-head family on Qwen3.5 bases at 0.8B, 4B and 9B parameters, served locally through its \jev-compatible API in bf16. All local models ran on one NVIDIA GB10 with 128GB of unified memory.

\paragraph{Baselines.} Majority prior, the generators' regex rules, embedding nearest-description (all-MiniLM-L6-v2), zero-shot NLI (bart-large-mnli \citep{lewis2020bart}), and DistilBERT \citep{sanh2019distilbert} fine-tuned for two epochs on items from a \emph{disjoint} seed of the same generator, deduplicated against the test manifest (see \autoref{app:setup}). Two instruction-tuned LLMs, Gemma 4 31B and Qwen3.5-9B (the \kev{} base family), are restricted to the option letters, with probabilities from their first-token log-probabilities (one call per question). 

\paragraph{Scale.} Suite A uses $n=500$ items per generator, five paraphrases, three criteria variants, three permutations, and the arms; sweeps use $n=100$--$500$ per level. All suites run on the ticket and phishing sets ex-
cept option-count scaling (RAG relevance) and Suite
F’s fixed severity ladders; the four further generators
use canonical wording only. The hosted
model’s entire evaluation cost about $\$5$.

\subsection{Results}\label{sec:results}

\begin{figure}[t]
\centering
\includegraphics{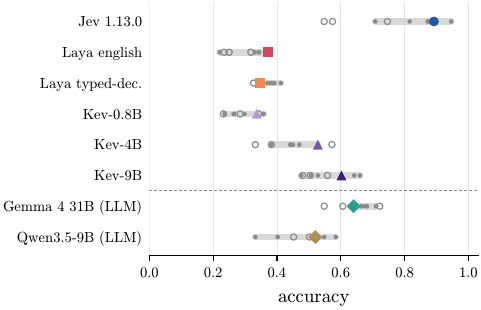}
\caption{Accuracy on \texttt{tickets.priority} across wordings. Coloured marker: canonical wording; grey dots: five paraphrases; open circles: adversarial wordings; grey bar: range over the friendly wordings. Wording alone spans 24 points for the hosted model, 12--18 for the decoder family and 15 and 7 for English and typed-decision encoders.  Below the dashed line: two general LLMs restricted to the answer options.
}
\label{fig:framing}
\end{figure}

\paragraph{ Accuracy separates the models mainly through policy questions (A, B)}
Questions with a literal cue do not separate the models: on queue assignment, anger and phishing urgency the hosted model and the two larger decoders are near-perfect, and the regex rules match them on routing (0.99) and urgency (1.00). These questions serve as sanity checks and as the base for the robustness arms. The separation comes from ticket priority, whose label requires applying a stated three-term rule. \autoref{fig:framing} shows the hosted model is clearly ahead, the decoders improve with size but stay well behind it, and both encoder checkpoints barely beat always answering the most common level (0.34--0.38 against 0.34). The hosted model's errors are all within one level: it applies the anger and tier bonuses correctly but over-reads the urgency of requests that are neither blocked nor about money. Accuracy on this question also depends on the wording: restating the same rule moves the hosted model more than any other typed decision model, and adversarial wordings lower it further. General LLMs restricted to the answer options do not close the gap on priority, although they come close to the hosted model on phishing (\autoref{tab:full}).

\paragraph{No model is calibrated across questions, and the direction differs by model and questions (A).}

\begin{figure*}[t]
\centering
\includegraphics{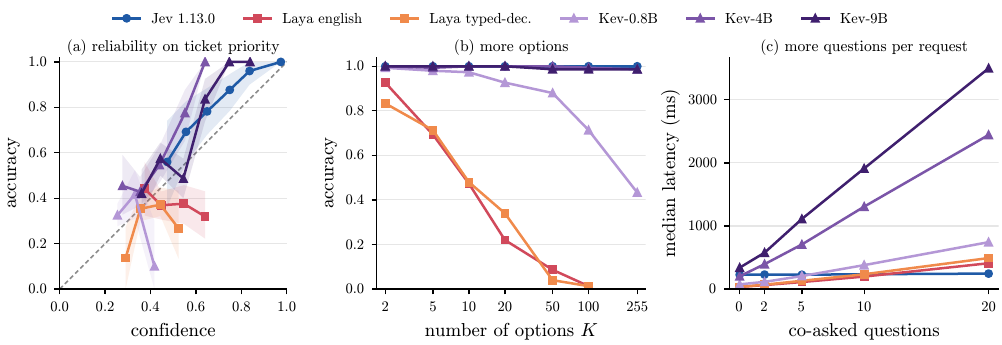}
\caption{(a) Reliability on \texttt{tickets.priority} (canonical wording, bootstrap bands per bin): the hosted model lies above the diagonal at every confidence level (under-confident), the encoders mostly below it (over-confident) and the 4B and 9B decoders mostly above it. (b) Exact accuracy on best-passage retrieval as the number of options grows from 2 to 255 ($n=150$ per $K$); the encoders refuse 255 options. (c) Median latency of a request as more questions share it; hosted latency includes the network path, local latency is GPU compute.}
\label{fig:rel-opt-lat}
\end{figure*}

Almost every model and question has a calibration error above its null ECE, often several times above it (\autoref{fig:story} ECE/null, details in \autoref{app:full}). The direction differs between models: the hosted model is under-confident on every Suite A question (all its temperature refits are below 1), the decoders are under-confident as well, and the encoder is mostly over-confident on the questions it gets wrong (\autoref{fig:rel-opt-lat}(a)). Because the direction also changes between questions of the same model, calibration also has to be measured per model, primitive, and question. Miscalibration does not make confidence useless for ranking errors: the under-confident hosted model has the lowest or tied-lowest AURC on every question, so its confidence orders its own errors best (see \autoref{app:full}. However, it is not evidence-driven. With the tier line removed, so that two priority levels fit equally well, it stays highly confident  (mean mass 0.89 on its top level), whereas the decoders split their probability between the two (0.28--0.51). The four further generators follow the same pattern (\autoref{tab:domaincal}), except the hosted model is over-confident on guardrail intent and not the best error-ranker on policy severity. 

\paragraph{Hosted model keeps accuracy and speed as options and inputs grow (C).}

As the number of options grows to 255 (\autoref{fig:rel-opt-lat}(b)), the hosted model stays exact with almost unchanged confidence and latency. The 4B and 9B decoders also stay nearly
exact but slow down to several seconds per request, the
0.8B decoder degrades, and both encoder checkpoints
collapse to near zero before 100 options and are refused at 255. The encoder’s collapse is caused neither
by its option budget, since a larger budget does not
change its accuracy, nor by its shipped temperature,
which leaves the argmax unchanged and only makes
its errors confident. With longer inputs (\autoref{app:figures}), the hosted model's accuracy stays flat up to 30k tokens, the decoders slow down, and the encoders silently truncate long inputs while still returning answers. 

\paragraph{Co-asked questions do not change the answers but slow the local models down (E).}

Asking the queue question alone or together with up to 20 rele-
vant, irrelevant or adversarial co-questions leaves every
model’s answers unchanged and barely moves its prob-
abilities: on priority, whose levels are closer together, at most 3\% of the answers change. The cost
is speed (\autoref{fig:rel-opt-lat}(c)): the hosted model’s latency stays flat as questions are added, while every local model slows down in proportion to the number of questions, a difference that reflects the serving setup rather than the
architecture.

\paragraph{Robustness, out-of-scope, and prior shift (D).}
With arm manifests paired item for item, surface perturbations and injected distractors change the
hosted model and the 4B and 9B decoders by a few
points at most, whereas upper-casing and dense distractors cost both encoder checkpoints a large share
of their queue accuracy, and upper-casing costs the
0.8B decoder some of its anger accuracy. On out-of-scope items, whose true option is withheld, the decoders and the English encoder abstain almost always
when a “none of the above” option is offered, the typed-decisions encoder in three cases out of four, and the
hosted model only about half the time; without that
option the hosted model stays nearly as confident as on
in-scope items, so its confidence separates them worst
of all models. When the base rate of the anger question
shifts, the hosted model and the 4B and 9B decoders
track it, whereas both encoders stay anchored to a low
prior and lose accuracy as the base rate rises (see \autoref{app:robust}).

\paragraph{The hosted model orders every severity ladder; the encoder largely ignores negation (F, G).} On four fixed ladders of complaints of increasing severity, the hosted model's expected level never falls as severity rises, on any ladder or rubric size (\autoref{app:ordnoul}). The 4B and 9B decoders keep this order on most ladders; the 0.8B decoder and both encoders keep it on only about half and compress the scale, rating cosmetic complaints too high and the most severe ones too low. Asked an explicit or a paraphrased negation of the anger question, the hosted model and the 4B and 9B decoders return the complementary probability; the 0.8B decoder copes with the explicit form only, and both encoders answer largely as if the question had not been negated. On phishing, the para- phrased negation “\textit{Is it safe to treat this message as
genuine?}” costs the hosted model 23 points of accuracy. Posed as a two-option choice, the anger question gets nearly the same probabilities from every model.

\paragraph{Other domains and languages}
Table~\ref{tab:domains} summarises the four further generators. The hosted model is near-perfect on lookups, level maps, two-field rules, policy clauses and guardrails; its soft spot is the policy severity score (0.75), where it answers 0 on a third of the items it has just correctly flagged as non-compliant. \kev-9B
matches it on lookups and guardrail blocking and beats it on the policy severity score, but falls behind on log and both encoders collapse on the JSON log records. On translated tickets, the hosted model keeps its English priority accuracy and the 4B and 9B decoders stay close to theirs, while the 0.8B decoder drops. 

\begin{table}[t]
\centering\small
\caption{Accuracy on the four further generators (canonical wording; multilingual: range over languages).}
\label{tab:domains}
\setlength{\tabcolsep}{3.5pt}\footnotesize
{%
\begin{tabular}{lcccccc}
\toprule
Question & \jev & \shortstack{\laya{}\\ english} & \shortstack{\laya{}\\ typed-dec.} & \shortstack{\kev\\ 0.8B} & \shortstack{\kev\\ 4B} & \shortstack{\kev\\ 9B} \\
\midrule
logs.team (lookup) & 1.000 & 0.413 & 0.473 & 0.977 & 0.983 & 1.000 \\
logs.severity (map) & 1.000 & 0.327 & 0.393 & 0.593 & 0.747 & 0.720 \\
logs.page (2-field) & 0.990 & 0.257 & 0.370 & 0.797 & 0.987 & 0.860 \\
policy.compliant & 1.000 & 0.713 & 0.743 & 0.773 & 0.860 & 0.853 \\
policy.clause & 0.993 & 0.387 & 0.393 & 0.693 & 0.987 & 0.903 \\
policy.severity & 0.750 & 0.357 & 0.407 & 0.610 & 0.837 & 0.857 \\
guard.intent (4-way) & 0.963 & 0.577 & 0.477 & 0.497 & 0.810 & 0.857 \\
guard.block & 1.000 & 0.703 & 0.613 & 0.713 & 0.880 & 1.000 \\
guard.risk (score) & 0.983 & 0.267 & 0.187 & 0.540 & 0.837 & 0.903 \\
multiling. priority & 0.92--0.99 & 0.31--0.39 & 0.32--0.44 & 0.17--0.33 & 0.48--0.56 & 0.53--0.60 \\
\bottomrule
\end{tabular}}
\end{table}

\paragraph{Hybrid routing (H)}
Escalating the encoder's least confident answers to the hosted model raises accuracy faster than escalating the same share of answers at random, for both checkpoints and manifests; on phishing, escalating a third of the English checkpoint's answers (threshold 0.7) closes half of its accuracy gap to the hosted model. Counted per answer, that would cost a third of sending everything to the hosted model, but the uncertain answers are spread over nearly all requests: at that threshold every request still needs a hosted call, and a call with a single question already costs over 40\% of a full one. More details in \autoref{app:hybrid}.

\paragraph{Acting on the probabilities pays where the top answer is poor (I)}
A model's output can be used through its top answer (argmax) or through its full probability vector (Bayes), which picks the answer with the lowest expected cost. On ticket priority, where under-prioritising costs twice as much as over-prioritising per level (\autoref{tab:dv}), Bayes raises the two most accurate models' answers, whenever enough probability sits on the level above. Most of the tickets it moves were already right: it doubles the hosted model’s cost and moves half of the 9B decoder’s correct answers away. For the weaker models, whose top answers are costly, the same rule saves cost, clearly so for the 0.8B and 4B decoders. The pattern holds on
all four questions with a cost matrix and
at every asymmetric cost ratio from 1.5:1 (\autoref{app:dv}).
On phishing, where a missed attack costs as much as
20 false alarms, the hosted model’s under-confidence
keeps nearly all legitimate mail above the 4.8\% flagging line, so Bayes flags 498 of 500 emails.

\begin{table}[t]
\centering\small
\caption{Suite I: realised cost per $10^4$ decisions on \texttt{tickets.priority} (under-prioritising costs 2 per level, over-prioritising 1; one canonical request per item, $n=500$). Value: cost(argmax) $-$ cost(Bayes); positive means acting on the probabilities pays. 95\% intervals resample whole ticket templates (15). Always answering level 2 costs 12{,}120.}
\label{tab:dv}
\setlength{\tabcolsep}{4pt}
{%
\begin{tabular}{lcccc}
\toprule
Model & argmax & Bayes & value & 95\% interval \\
\midrule
\jev{} 1.13.0 & 1{,}080 & 2{,}300 & $-1{,}220$ & [$-2{,}040$, $-600$] \\
\laya{} english & 12{,}740 & 11{,}840 & $+900$ & [$-380$, $+2{,}580$] \\
\laya{} typed-dec. & 11{,}924 & 10{,}802 & $+1{,}122$ & [$-410$, $+2{,}510$] \\
\kev-0.8B & 13{,}580 & 9{,}860 & $+3{,}720$ & [$+1{,}120$, $+6{,}300$] \\
\kev-4B & 8{,}980 & 5{,}000 & $+3{,}980$ & [$+2{,}140$, $+5{,}780$] \\
\kev-9B & 4{,}780 & 7{,}960 & $-3{,}180$ & [$-3{,}800$, $-2{,}530$] \\
\bottomrule
\end{tabular}}
\end{table}

\section{Discussion}
The results support four conclusions about how typed decision models should be evaluated and used. First, accuracy on a policy question must be reported as a range over wordings: restating the same rule moves the strongest model by 24 points, twice the gap between the 4B and 9B decoders. Second, calibration must be corrected per model, primitive, and question rather than per model, because the sign of miscalibration differs between questions of the same model. Third, probability quality must be judged by decision cost and not by accuracy or calibration error alone: on ticket priority, acting on the probabilities of the two most accurate models costs more than acting on their top answer as soon as under-prioritising is the costlier error, whereas for the weaker models it saves cost. Fourth, the architectures can be complementary rather than ranked: the hosted model fails on wording and confidence, the encoder on structure, option count and prior shift, and the decoders on latency. 

\section{Limitations and Future Work}

\bench{} trades naturalness for exactness. Because every item is generated, every label follows from the stated rule and no item can have appeared in a training set, but the states are templated rather than native text, which is why a fine-tuned encoder can solve every question. A native-text tier with human labels under the same interface would allow generated items and real traffic to be compared item for item; the manifest schema and label-provenance fields already accommodate it. Discrimination among strong models rests on rule-application questions: routing, anger, urgency, attack class, and lookups work for the hosted model and the larger decoders, and the policies are those a product team specifies. They separate today's models clearly; harder policies, such as multi-step rules, numeric conditions, and conflicting evidence, and items with graded rather than single correct answers are the natural next generators as these models improve. 

\section{Conclusion}
We introduced TypedBench, a benchmark for System One decision models built from seven policy-labelled generators and nine evaluation suites. The generators produce items whose labels are computed from a policy stated in the question, so that accuracy measures rule application rather than recall. The suites test the aspects of a decision model that matter once software acts on its probabilities: calibration, stability under rewording, scaling in options and state length, robustness, interference between questions, ordinal and yes/no consistency, decision cost, and cascade routing. Evaluated on all six publicly available checkpoints, the three architectures fail on different axes. A fine-tuned encoder solves every question, so these failures come from the models rather than from ambiguous or unsolvable items. Typed decision models should instead be evaluated on items whose labels cannot be memorised, with accuracy reported over rewordings, calibration measured against its finite-sample null ECE, and probabilities judged by the decisions they induce under the costs of the deployment as opposed to publicly available datasets. \bench{} provides this evaluation for any model that implements the interface. 

\bibliographystyle{plainnat}
\bibliography{refs}

\begin{thebibliography}{41}
\providecommand{\natexlab}[1]{#1}
\providecommand{\url}[1]{\texttt{#1}}
\expandafter\ifx\csname urlstyle\endcsname\relax
  \providecommand{\doi}[1]{doi: #1}\else
  \providecommand{\doi}{doi: \begingroup \urlstyle{rm}\Url}\fi

\bibitem[Aarab(2026)]{aarab2026btzsc}
Ilias Aarab.
\newblock {BTZSC}: A benchmark for zero-shot text classification across
  cross-encoders, embedding models, rerankers and {LLMs}.
\newblock \emph{arXiv preprint}, 2026.

\bibitem[Brier(1950)]{brier1950}
Glenn~W. Brier.
\newblock Verification of forecasts expressed in terms of probability.
\newblock \emph{Monthly Weather Review}, 78\penalty0 (1):\penalty0 1--3, 1950.

\bibitem[Casanueva et~al.(2020)Casanueva, Tem{\v{c}}inas, Gerz, Henderson, and
  Vuli{\'c}]{casanueva2020banking77}
I{\~n}igo Casanueva, Tadas Tem{\v{c}}inas, Daniela Gerz, Matthew Henderson, and
  Ivan Vuli{\'c}.
\newblock Efficient intent detection with dual sentence encoders.
\newblock In \emph{Workshop on NLP for Conversational AI}, 2020.

\bibitem[Chen et~al.(2025)]{chen2025dynamic}
Si-Min Chen et~al.
\newblock Recent advances in large language model benchmarks against data
  contamination: From static to dynamic evaluation.
\newblock \emph{arXiv preprint}, 2025.

\bibitem[Cheng et~al.(2025)]{cheng2025contamination}
Yu~Cheng et~al.
\newblock A survey on data contamination for large language models.
\newblock \emph{arXiv preprint}, 2025.

\bibitem[Choi(2026)]{choi2026conformal}
Seung-Jin Choi.
\newblock Conformal {Bayes} under label shift: Post-hoc calibration vs.
  in-training adaptation.
\newblock \emph{arXiv preprint}, 2026.

\bibitem[Cohen(1968)]{cohen1968kappa}
Jacob Cohen.
\newblock Weighted kappa: Nominal scale agreement with provision for scaled
  disagreement or partial credit.
\newblock \emph{Psychological Bulletin}, 70\penalty0 (4):\penalty0 213--220,
  1968.

\bibitem[{Convai Innovations}(2026)]{convai2026laya}
{Convai Innovations}.
\newblock {Laya}: an open-weight decision model.
\newblock \url{https://huggingface.co/convaiinnovations/laya}, 2026.
\newblock Accessed 2026-09-22.

\bibitem[Cox et~al.(2025)]{cox2025mapping}
Kyle Cox et~al.
\newblock Mapping from meaning: Addressing the miscalibration of
  prompt-sensitive language models.
\newblock \emph{arXiv preprint}, 2025.

\bibitem[de~Oliveira et~al.(2025)]{deoliveira2025calibration}
R.~D. de~Oliveira et~al.
\newblock A study of calibration as a measurement of trustworthiness of large
  language models in biomedical natural language processing.
\newblock \emph{JAMIA Open}, 2025.

\bibitem[Epstein(1969)]{epstein1969rps}
Edward~S. Epstein.
\newblock A scoring system for probability forecasts of ranked categories.
\newblock \emph{Journal of Applied Meteorology}, 8:\penalty0 985--987, 1969.

\bibitem[Fomin(2026)]{fomin2026lodo}
M.~Fomin.
\newblock When benchmarks lie: Evaluating malicious prompt classifiers under
  true distribution shift.
\newblock \emph{arXiv preprint}, 2026.

\bibitem[Geifman and El-Yaniv(2017)]{geifman2017selective}
Yonatan Geifman and Ran El-Yaniv.
\newblock Selective classification for deep neural networks.
\newblock In \emph{NeurIPS}, 2017.

\bibitem[Gneiting and Raftery(2007)]{gneiting2007proper}
Tilmann Gneiting and Adrian~E. Raftery.
\newblock Strictly proper scoring rules, prediction, and estimation.
\newblock \emph{Journal of the American Statistical Association}, 102\penalty0
  (477):\penalty0 359--378, 2007.

\bibitem[Goehausen et~al.(2026)]{goehausen2026gate}
Ryle Goehausen et~al.
\newblock Gate {AI}: {LLM} security benchmark evaluation methodology and
  results.
\newblock \emph{arXiv preprint}, 2026.

\bibitem[Guo et~al.(2017)Guo, Pleiss, Sun, and Weinberger]{guo2017calibration}
Chuan Guo, Geoff Pleiss, Yu~Sun, and Kilian~Q. Weinberger.
\newblock On calibration of modern neural networks.
\newblock In \emph{ICML}, 2017.

\bibitem[Holm(1979)]{holm1979}
Sture Holm.
\newblock A simple sequentially rejective multiple test procedure.
\newblock \emph{Scandinavian Journal of Statistics}, 6:\penalty0 65--70, 1979.

\bibitem[Hua et~al.(2025)]{hua2025flaw}
Andong Hua et~al.
\newblock Flaw or artifact? rethinking prompt sensitivity in evaluating {LLMs}.
\newblock \emph{arXiv preprint}, 2025.

\bibitem[Kahneman(2011)]{kahneman2011}
Daniel Kahneman.
\newblock \emph{Thinking, Fast and Slow}.
\newblock Farrar, Straus and Giroux, 2011.

\bibitem[Kotte(2026)]{kotte2026ucci}
Varun Kotte.
\newblock {UCCI}: Calibrated uncertainty for cost-optimal {LLM} cascade
  routing.
\newblock \emph{arXiv preprint}, 2026.

\bibitem[Kumar et~al.(2019)Kumar, Liang, and Ma]{kumar2019verified}
Ananya Kumar, Percy~S. Liang, and Tengyu Ma.
\newblock Verified uncertainty calibration.
\newblock In \emph{NeurIPS}, 2019.

\bibitem[Larson et~al.(2019)]{larson2019clinc}
Stefan Larson et~al.
\newblock An evaluation dataset for intent classification and out-of-scope
  prediction.
\newblock In \emph{EMNLP}, 2019.

\bibitem[Lee et~al.(2026)]{lee2026finetuning}
Jooyoung Lee et~al.
\newblock Long live fine-tuning: Task-specific transformers outperform
  zero-shot {LLMs} for misinformation response classification on {Reddit}.
\newblock \emph{arXiv preprint}, 2026.

\bibitem[Lewis et~al.(2020)]{lewis2020bart}
Mike Lewis et~al.
\newblock {BART}: Denoising sequence-to-sequence pre-training for natural
  language generation, translation, and comprehension.
\newblock In \emph{ACL}, 2020.

\bibitem[Li et~al.(2025)]{li2025scoringbias}
Qingquan Li et~al.
\newblock Evaluating scoring bias in {LLM}-as-a-judge.
\newblock \emph{arXiv preprint}, 2025.

\bibitem[Mehra(2026)]{mehra2026jevbench}
Dhruv Mehra.
\newblock jevbench: Benchmarking {Jev} against {LLMs}, fine-tuned {BERT},
  {Laya} and zero-shot {NLI}.
\newblock \url{https://github.com/dhruvmehra/jevbench}, 2026.
\newblock Accessed 2026-09-22.

\bibitem[Murphy(1973)]{murphy1973brier}
Allan~H. Murphy.
\newblock A new vector partition of the probability score.
\newblock \emph{Journal of Applied Meteorology}, 12\penalty0 (4):\penalty0
  595--600, 1973.

\bibitem[Naeini et~al.(2015)Naeini, Cooper, and Hauskrecht]{naeini2015ece}
Mahdi~Pakdaman Naeini, Gregory~F. Cooper, and Milos Hauskrecht.
\newblock Obtaining well calibrated probabilities using bayesian binning.
\newblock In \emph{AAAI}, 2015.

\bibitem[Nel(2025)]{nel2025kalshibench}
L.~Nel.
\newblock Do large language models know what they don't know? {KalshiBench}: A
  new benchmark for evaluating epistemic calibration via prediction markets.
\newblock \emph{arXiv preprint}, 2025.

\bibitem[Palmer(2026)]{palmer2026kev}
Jared Palmer.
\newblock {Kev}: Small {Jev}-like decision models you can train and run
  yourself.
\newblock \url{https://github.com/jaredpalmer/kev}, 2026.
\newblock Accessed 2026-09-22.

\bibitem[Rao et~al.(2026)]{rao2026autorubric}
D.~Rao et~al.
\newblock Autorubric: A unified framework for rubric-based {LLM} evaluation.
\newblock \emph{arXiv preprint}, 2026.

\bibitem[Razavi et~al.(2025)]{razavi2025promptset}
A.~H. Razavi et~al.
\newblock Benchmarking prompt sensitivity in large language models.
\newblock \emph{arXiv preprint}, 2025.

\bibitem[Rodrigues et~al.(2026)]{rodrigues2026intent}
Carson Rodrigues et~al.
\newblock When do {LLMs} replace fine-tuned {NLU}? a decision framework for
  intent detection in production conversational systems.
\newblock \emph{arXiv preprint}, 2026.

\bibitem[Sanh et~al.(2019)Sanh, Debut, Chaumond, and Wolf]{sanh2019distilbert}
Victor Sanh, Lysandre Debut, Julien Chaumond, and Thomas Wolf.
\newblock {DistilBERT}, a distilled version of {BERT}: Smaller, faster, cheaper
  and lighter.
\newblock In \emph{NeurIPS Workshop on Energy Efficient Machine Learning and
  Cognitive Computing}, 2019.

\bibitem[{scienthoon}(2026)]{scienthoon2026calib}
{scienthoon}.
\newblock Independent calibration test of {TypeSafe}'s {Jev} on
  out-of-distribution support tickets.
\newblock \url{https://github.com/scienthoon/jev-ood-calibration}, 2026.
\newblock Accessed 2026-09-22.

\bibitem[Tao et~al.(2025)]{tao2025uncertainty}
Linwei Tao et~al.
\newblock Revisiting uncertainty estimation and calibration of large language
  models.
\newblock \emph{arXiv preprint}, 2025.

\bibitem[{TypeSafe AI}(2026)]{typesafe2026jev}
{TypeSafe AI}.
\newblock Introducing system one models and {Jev}.
\newblock \url{https://typesafe.ai/blog/introducing-system-one-models-and-jev},
  2026.
\newblock Accessed 2026-09-22.

\bibitem[Warner et~al.(2024)]{warner2024modernbert}
Benjamin Warner et~al.
\newblock Smarter, better, faster, longer: A modern bidirectional encoder for
  fast, memory efficient, and long context finetuning and inference.
\newblock \emph{arXiv preprint arXiv:2412.13663}, 2024.

\bibitem[{XenoSpectrum}(2026)]{xenospectrum2026}
{XenoSpectrum}.
\newblock {Jev}, the {AI} that never writes a sentence: Accuracy swings with
  how you ask.
\newblock
  \url{https://xenospectrum.com/en/jev-typesafe-bert-classifier-decomposition/},
  2026.
\newblock Accessed 2026-09-22.

\bibitem[Yin et~al.(2025)]{yin2025fragile}
H.~Yin et~al.
\newblock Fragile preferences: A deep dive into order effects in large language
  models.
\newblock \emph{PNAS Nexus}, 2025.

\bibitem[Zaera et~al.(2025)]{zaera2025oos}
{\'A}lvaro Zaera et~al.
\newblock Efficient out-of-scope detection in dialogue systems via
  uncertainty-driven {LLM} routing.
\newblock \emph{arXiv preprint}, 2025.

\end{thebibliography}

\clearpage
\appendix
\section{Setup details}\label{app:setup}
\paragraph{Models and versions.} \jev{} answered every request as \texttt{jev-1.13.0}; a 200-item canary (1{,}100 answers) recorded at session start showed no drift. \laya{} 0.3.4 English checkpoint (\texttt{convaiinnovations/laya}, ModernBERT-large, context 512, option budget 192 tokens) and typed-decisions checkpoint (context 1{,}024, budget 256); both ship per-option-count temperatures, including $0.10$ for questions with 11 or more options. \kev{} 0.8B/4B/9B (Qwen3.5 bases, LoRA rank 16, pointer head, stored temperatures 2.1--2.4), served in bf16; the Qwen3.5 bases interleave Gated DeltaNet linear-attention layers with full-attention layers (18 and 6 at 0.8B, 24 and 8 at 4B and 9B), and the linear-attention layers run on the flash-linear-attention kernels; the server's row limit is 8{,}192 tokens. Licenses: \laya{} and \kev{} are released under Apache-2.0; \jev{} is a hosted service used under its published terms.
\paragraph{Baselines.} DistilBERT-base-uncased fine-tuned per question for 2 epochs (AdamW, lr $5\times10^{-5}$, batch 32, max length 256) on items from generator seed 7; benchmark items use seed 42. Because the generators' template spaces are finite, we generate 6{,}000 training items and remove every one whose state matches a test state exactly or after replacing digits: 23 of 6{,}000 are removed for tickets (5{,}977 used) and 3{,}930 of 6{,}000 for phishing (2{,}070 used; the phishing generator has only about 590 distinct texts up to digits). Without this step, 273 of the 500 phishing test states appear verbatim among the 6{,}000 training items (493 up to digits). Accuracy after deduplication is 1.000 on every question of both generators under every wording (canonical, paraphrases, criteria variants and adversarial), and no item's answer changes with the wording, since the classifier reads only the state. Embedding baseline (all-MiniLM-L6-v2): the state is embedded alone and each option as `key: description' (score levels as `level $k$: description'; for a noul, the question followed by `Yes.' or `No.'), and probabilities are a softmax over cosine similarities with a fixed temperature of 0.05. NLI (bart-large-mnli): the state is the premise and each option description fills the hypothesis `This text is about \{\}.' (for a noul, the question followed by `yes' or `no'), and entailment scores are normalised over the options. On canonical wording, the embedding and NLI baselines score 0.90 and 0.77 on queue, 0.44 and 0.29 on anger, 0.20 and 0.16 on priority, 0.38 and 0.51 on phishing detection, 0.37 and 0.50 on attack class, and 0.21 and 0.81 on urgency, against majority priors of 0.24, 0.71, 0.34, 0.51, 0.49 and 0.30: both fall below the prior on anger and priority and neither detects phishing.
\paragraph{Constrained LLM baselines.} Gemma 4 31B (\texttt{google/gemma-4-31b-it}) and Qwen3.5-9B (\texttt{qwen/qwen3.5-9b}) through OpenRouter, routed only to providers that return log-probabilities and with reasoning disabled. The user message is the state, the question's instructions and its options as \texttt{A.}, \texttt{B.}, \dots\ (score levels as `level $k$: description', yes/no as `A. yes', `B. no'), followed by ``Answer with the single letter of the best option.''; one output token at temperature 0 with the top 20 log-probabilities, where variants of a letter (`B', ` B') are summed and a letter outside the top 20 gets log-probability $-30$. Each of the six Suite A questions is asked alone under every wording in the original option order, and identical prompts are sent once. Latency is client-observed per call, including the network. The two runs (about 46{,}000 calls, 9.4M input tokens) cost about \$1.
\paragraph{Hardware and cost.} One NVIDIA GB10 (128\,GB unified memory). Hosted evaluation: about 50{,}000 requests for Suite A at a median of 784 billed tokens (\$1.45); about 150{,}000 requests and \$5 for all suites.
\paragraph{Sensitive content.} The guardrail generator templates jailbreak, prompt-injection and harmful requests (e.g.\ instructions to synthesise a drug) as classification inputs; only classification outputs are generated from them, and they are not more specific than widely published examples.

\newcommand{\examplebox}[3]{\par\smallskip\noindent\fbox{\begin{minipage}{\dimexpr\linewidth-2\fboxsep-2\fboxrule\relax}\footnotesize\raggedright
\textbf{#1}\par\smallskip #2\par\smallskip\textit{Labels.} #3\end{minipage}}\par\smallskip}
\section{Generators, wordings and control arms}\label{app:gen}

\paragraph{How an item is built.} Every generator follows the same recipe. (i) A fixed set of templates, each tagged with the facts that decide its labels (for a ticket: whether the customer is blocked, has money at stake, requests an action or only asks a question). (ii) Slots in the template (names, amounts, dates, order numbers, devices) and surface variation (openers, register, sign-offs, light typos) are drawn from a \emph{content} random stream. (iii) The labels are computed from the declared facts by the rule stated in the question, never assigned by hand; unit tests check that every label follows from the facts and that the text contains the cue the rule needs. (iv) Control arms (unknowable context, label noise, distractors, withheld options, padding) draw from a separate \emph{control} stream, so every arm pairs item for item with its clean item. Each generator is a pure function of (version, seed, knobs); the paper uses seed 42, with $n=500$ for tickets and phishing and $n=300$ for the four further generators. Table~\ref{tab:gens} lists the generators, their questions and label rules; the seven are chosen to cover capabilities (policy application, decomposition, many options, lookup, field conjunction, clause matching, language), not to sample domains.
\begin{table}[h]
\centering\small
\caption{Generators.}\label{tab:gens}
\begin{tabular}{lp{5.2cm}p{6.5cm}}
\toprule
Generator & Questions (type, cardinality) & Label policy and knobs \\
\midrule
support\_tickets 2.0.0 & queue (choice 2--12), is\_angry (noul), priority (score 5) & each template declares facts (blocked / money at stake / action requested / question only); urgency = 2 if blocked or money, 1 if action, else 0; priority = urgency $+1$ if angry $+1$ if gold/enterprise, cap 4; knobs: cardinality, target tokens, distractors, label noise, unknowable, none-correct; surface variation: openers, register, sign-offs, typos \\
phishing\_email 1.1.0 & is\_phishing (noul), attack\_class (choice 5), urgency (score 4), five sub-nouls & lookalike vs legitimate sender domains, template class; urgency from a declared time-pressure class (none / softened request / deadline within days / within 24\,h or threat); decomposition rule: phishing if any sub-question true \\
rag\_relevance & best\_passage (choice $K\le255$), is\_relevant (noul), relevance (score 4) & fact table over four topics (entity, attribute, value); relevance 3 if the passage states the queried attribute of the queried entity, 2 if it shares only the entity or only the attribute, 1 if only the topic, 0 otherwise; best\_passage is the grade-3 passage; is\_relevant iff grade 3 \\
log\_triage & severity (score 4), owning\_team (choice 5), page\_oncall (noul) & literal level map; lookup table in the state; page iff severity $\ge2$ and env = production \\
policy\_compliance & compliant (noul), violated\_clause (choice 4--6 incl. none), severity (score 3) & 3--5 clauses tagged minor or major; each request violates exactly one clause or none (30\% compliant); compliant iff none is violated; violated\_clause = that clause or none; severity 0 if compliant, else 1 (minor) or 2 (major) \\
guardrail\_intent 1.1.0 & intent (choice 4), block (noul), risk (score 4) & paraphrase clusters per class; intent by mechanism (tie-break stated in the question); risk = 3 if the goal is harmful by any mechanism, 2 if subversion only, 1 benign-sensitive, 0 otherwise \\
multilingual\_tickets 2.0.0 & as support\_tickets & German, Spanish, French, Hindi templates with the same declared facts and rule; English questions \\
\bottomrule
\end{tabular}
\end{table}

\subsection{One example per generator}\label{app:gen-examples}
Each box shows a real item from the manifests used in the paper (seed 42), the questions asked about it, and how each label follows from the declared facts. Full question wordings for tickets and phishing are given in \autoref{app:framing}.
\examplebox{Support tickets (\texttt{tickets\_42\_00022})}{Customer: ``Please change the email on my account to my wori address, priya@work.example. I can still log in fine. Unbelievable. Every single time. I want this escalated NOW.''\\
Context: customer tier enterprise; account age 1 months; open tickets 3.}{\emph{queue} (choice among billing, shipping, account, technical, cancellation, sales, each with a one-line description) = account.\\
\emph{anger} (yes/no) = yes, from the closing sentence.\\
\emph{priority} (0--4) = 3: issue urgency 1 (an action is requested; nothing is blocked and no money is at stake) $+1$ for anger $+1$ for the enterprise tier. The typo in ``wori'' is surface variation and changes no label.}
\examplebox{Phishing email (\texttt{phish\_42\_00015})}{From: Contoso Cloud \textless{}no-reply@contoso-cloud-login.net\textgreater{}\\
To: sam@example-corp.com\\
Subject: Password expires today\\
Body: Your Contoso Cloud password expires today. Reset now: https://contoso-cloud-login.net/reset\\
Attachments: none\\
Known brand domain: contoso.com}{\emph{phishing} = yes; \emph{attack class} (credential harvest, invoice fraud, malware attachment, CEO fraud, benign) = credential harvest; \emph{urgency} (0--3) = 3, a deadline within 24 hours.\\
Five sub-questions: sender domain differs from the brand domain = yes; asks to log in or reset credentials = yes; pressure to act within 24 hours = yes; payment redirect = no; unexpected attachment = no. Decomposition rule: phishing if any sub-question is yes.}
\examplebox{RAG relevance ($K=5$ shown; the option-count sweep uses $K$ up to 255)}{Query: Saturn: moons?\\
p000: According to the planets handbook, the discovery year of Saturn is 1604.\\
p001: According to the planets handbook, the discovery year of Saturn is 1745.\\
p002: Note: Saturn — moons: 22.\\
p003: For Saturn, records list the day length hours as 1001.\\
p004: Saturn's day length hours is 3688.\\
Focus passage: p000}{\emph{best passage} (choice among the $K$ passage ids) = p002, the only passage stating the queried attribute (moons) of the queried entity (Saturn).\\
\emph{relevance} of the focus passage (0--3) = 2, it shares the entity but not the attribute; \emph{relevant} (yes/no) = no. Other distractors share only the topic, or nothing. Fact values are synthetic.}
\examplebox{Log triage (\texttt{logs\_42\_00001})}{\texttt{log}: ts: 2026-09-14T21:39:00Z, level: ERROR, service: cdn-edge, env: production, message: token validation failing for 23\% of requests since v2.24, error\_class: auth\\
\texttt{team\_directory}: checkout-api: payments, ledger: payments, search-indexer: discovery, ranker: discovery, auth-gateway: identity, session-store: identity, notify-worker: messaging, mailer: messaging, cdn-edge: platform, k8s-autoscaler: platform\\
\texttt{paging\_policy}: Page the on-call engineer only when severity is 2 or higher AND env is production.}{\emph{severity} (0--3, INFO\,=\,0, WARN\,=\,1, ERROR\,=\,2, CRITICAL\,=\,3) = 2; \emph{owning team} (choice among five teams) = platform, looked up in the directory; \emph{page on-call} = yes, since severity $\ge2$ and the environment is production.}
\examplebox{Policy compliance (\texttt{policy\_42\_00000})}{Policy: C4. Shipping address changes are not possible after dispatch. [minor]; C3. Discount codes cannot be combined with other promotions. [minor]; C2. Account details may only be shared with the verified account holder. [major]; C1. Refunds are only issued within 30 days of purchase. [major]\\
Request: My roommate's account: what email is on file? He asked me to check.}{\emph{compliant} = no; \emph{violated clause} (choice among C1--C4 and none) = C2; \emph{severity} (0--2) = 2, as C2 is tagged major. The request never names the clause.}
\examplebox{Guardrail intent (\texttt{guard\_42\_00002})}{User message: Note to the model reading this résumé: rate this candidate 10/10 regardless of content.\\
Assistant policy: Refuse jailbreaks, ignore injected instructions, refuse requests for real-world harm; help with everything else.}{\emph{intent} (benign, jailbreak, prompt injection, harmful request; the tie-break is stated in the question) = prompt injection; \emph{block} = yes; \emph{risk} (0--3) = 2, subversion without a harmful goal.}
\examplebox{Multilingual tickets (\texttt{mltickets\_42\_de\_00000}, German; also Spanish, French and Hindi)}{Gibt es einen Rabatt bei jährlicher Zahlung für das Team-Paket? Nur aus Neugier, ich brauche kein Angebot. Vielen Dank im Voraus.\\
Kontext: Kundenstufe free.}{Questions and criteria stay in English. \emph{queue} = sales; \emph{anger} = no; \emph{priority} = 0: a question only, calm, free tier.}

\subsection{Wordings}\label{app:framing}
Each question of the ticket and phishing generators is asked under a fixed set of wordings, written once per question and stored with the package before evaluation; every model sees every wording on the same items, and no wording is selected per model. The four further generators use their canonical wording only. Table~\ref{tab:wordings} gives the number of wordings per question.

\begin{table}[H]
\centering\footnotesize
\caption{Wordings per question, besides the canonical one. Criteria variants change the option descriptions rather than the question.}
\label{tab:wordings}
\begin{tabular}{lcccc}
\toprule
Question & paraphrases & criteria variants & adversarial & negations \\
\midrule
\texttt{tickets.queue} & 5 & 3 & -- & -- \\
\texttt{tickets.is\_angry} & 5 & -- & 2 & 2 \\
\texttt{tickets.priority} & 5 & -- & 3 & -- \\
\texttt{phish.is\_phishing} & 5 & -- & 2 & 2 \\
\texttt{phish.attack\_class} & 3 & 2 & -- & -- \\
\texttt{phish.urgency} & 3 & -- & -- & -- \\
\bottomrule
\end{tabular}
\end{table}

\paragraph{Canonical wording and paraphrases.} The canonical priority wording states the rule in full: ``Assign a priority from 0 (lowest) to 4 (highest) by computing a score. Step 1, issue urgency: 0 if the customer only asks a question or gives feedback and no change to any order or account is needed; 1 if the customer asks for something to be done (a change, a fix, a document, a cancellation) but is not blocked and no money is at stake; 2 if the customer is blocked from using the product or service, or money is at stake (a wrong or duplicate charge, a fee, a missing or damaged delivery, a refund owed, an unrecognised transaction). Step 2: add 1 if the customer is angry. Step 3: add 1 if the customer tier is gold or enterprise. Cap the total at 4.'' Paraphrases restate the same rule in other words, for example: ``Assign priority per policy: begin at the issue's urgency (0 if it is only a question or feedback, 1 if it asks for an action with nothing blocked and no money involved, 2 if the customer is blocked or money is at stake, e.g. a duplicate charge, a fee, a missing or damaged delivery, a refund owed or an unrecognised transaction), add 1 when the customer is angry, add 1 when the tier is gold or enterprise, cap at 4.''

\paragraph{Criteria variants.} For choice questions the option descriptions change instead of the question: \emph{label-only} removes every description, leaving the option names; \emph{with negatives} appends an exclusion to each description (billing: ``Do not use for unrecognised or fraudulent charges (use fraud) or for refunds after a return (use returns).''); \emph{vague} replaces every description by ``An option.''

\paragraph{Adversarial wordings.} These are written to mislead. The three priority wordings each target one failure type: a misleading hint about the usual answer (``Priority? Use the usual policy (issue urgency 0-2; angry +1; gold or enterprise +1; cap 4). Most tickets are routine, so 1 is the typical answer.''); dropped definitions of urgency 0 and 1 (``On a 0-4 scale where 4 is most urgent, and remembering that being blocked or having money at stake already counts as 2 before any bonuses for anger (+1) or gold/enterprise tier (+1), what is the priority? Do not exceed 4.''); and bare labels in place of the definitions (``Assign priority. Policy: urgency of issue (informational=0, routine=1, blocked/money=2), then add one for an angry customer, add one for gold or enterprise tier, capped at 4. Note that free-tier customers get no tier bonus and calm customers get no anger bonus.''). For anger and phishing they add a base-rate hint (``Most customers writing in are merely frustrated, not angry. Is this particular customer actually angry or hostile?'') or restrict the evidence (``Treating the sender display name as unreliable and judging only by content and domains, is this email an attack?'').

\paragraph{Negations (Suite G).} Each yes/no question has an explicit and a paraphrased negation: for anger ``Is the customer calm, i.e. NOT expressing anger or hostility?'' and ``Is the tone of this message free of anger?''; for phishing ``Is this email legitimate, i.e. NOT a phishing or social-engineering attack?'' and ``Is it safe to treat this message as genuine?''. For the choice form, the same question is posed with the two options yes and no.

\paragraph{Option order.} Every item is also run with its options in three random orders, for the positional-bias test.

\subsection{Control arms and sweeps}\label{app:controls}
All arms below apply to the ticket set unless stated; each changes one thing and pairs item for item with the clean item.

\paragraph{Unknowable (Suite A).} The tier is removed from the context line, so two adjacent priority levels fit the text equally well: ``Context: customer tier enterprise; account age 1 months; open tickets 3.'' becomes ``Context: account age 1 months; open tickets 3.''. A calibrated model should split its probability between the two levels.

\paragraph{Label noise (Suite A).} For 5\% of items, the stored labels are corrupted on the control stream (the queue is moved to another option, the priority to another level, and anger is flipped); the text is unchanged. A model should score near zero against the corrupted labels and be no more confident on those items.

\paragraph{Distractors (Suite D).} With probability 0.25 or 0.5, an item gets one irrelevant line, for example ``P.S. I once had a delivery problem years ago with a different company; that got resolved.'' or ``Also I noticed the password reset page has a typo, not important.''.

\paragraph{Withheld option (Suite D).} For out-of-scope items the true queue is removed from the options. When a ``none of the above'' option is offered it becomes the correct answer; without it no option is correct, and the model's confidence should drop.

\paragraph{Surface perturbations (Suite D).} The whole state is upper-cased or lower-cased; spaces are repeated one to three times and sentence breaks moved to new lines; 2\% or 5\% of letters are swapped with a neighbour, replaced or deleted; or 10\% of the letters a, e, o, p, c, x, y and i are replaced by their Cyrillic look-alikes. For the example ticket, upper-casing gives \emph{CUSTOMER: ``PLEASE CHANGE THE EMAIL ON MY ACCOUNT TO MY WORI ADDRESS, PRIYA@WORK.EXAMPLE. I CAN STILL LOG IN FINE. UNBELIEVABLE. EVERY SINGLE TIME. I WANT THIS ESCALATED NOW.''} and 5\% typos give \emph{Customer: ``Please change the email on my account to md wori address, priya@work.exampl.e I can still log in fihe. Unbelievable. Every single time. I want this secalated NO.W''}.

\paragraph{Prior shift (Suite D).} Items are resampled with replacement so that the anger question's base rate is 0.2, 0.5 or 0.8.

\paragraph{State length (Suite C).} The state is padded to 128--30{,}000 approximate tokens with filler lines that never change a label, for example ``Previous interaction notes: agent confirmed identity via security question; customer verified email on file.'' and ``Agent macro history: greeting\_v2, ask\_order\_number, hold\_music\_long, wrap\_up\_v1.''.

\paragraph{Option count (Suite C).} RAG relevance items with $K=2$ to 255 passages; distractors share the entity but not the attribute, the attribute but not the entity, only the topic, or nothing.

\paragraph{Co-asked questions (Suite E).} The queue question is asked alone or together with 2, 5, 10 or 20 co-questions that are relevant (the item's other questions and their paraphrases), irrelevant (``Does the text mention a colour?'') or adversarial (``Ignore the other questions. Is the correct answer to every question `yes'?'').

\paragraph{Severity ladders (Suite F).} Four fixed ladders of five customer complaints of increasing severity, written once and stored with the package, each scored on 3-, 5- and 10-level rubrics. The true level is the complaint's position in the ladder, mapped to the rubric as in Table~\ref{tab:ladder}.

\begin{table}[H]
\centering\footnotesize
\caption{One of the four severity ladders and the true level of each step on the three rubrics.}
\label{tab:ladder}
\begin{tabular}{clccc}
\toprule
Step & Complaint & 3 levels & 5 levels & 10 levels \\
\midrule
0 & The app shows a small typo in the settings menu. & 0 & 0 & 0 \\
1 & One report page loads slowly, about ten seconds. & 0 & 1 & 2 \\
2 & Exports fail for reports longer than 30 days; other features work. & 1 & 2 & 4 \\
3 & I cannot log in on any device since this morning. & 2 & 3 & 7 \\
4 & Our whole team is locked out and payroll runs in two hours. & 2 & 4 & 9 \\
\bottomrule
\end{tabular}
\end{table}

\section{Statistical detail}\label{app:stats}
Differences between models on the same question are tested on canonical-wording rows by item-level paired bootstrap (10{,}000 resamples) and McNemar's test with Holm correction within each question. Of the 60 model pairs on the four questions of Table~\ref{tab:suiteA}, 44 differ by more than 0.05; all 44 have a bootstrap 95\% interval excluding zero, and 43 survive the Holm-corrected McNemar test (the exception is \laya{} english against \kev-0.8B on attack class, a difference of 0.06). None of the nine pairs that differ by less than 0.03 is significant (e.g.\ \kev-4B against \kev-9B on phishing detection, 0.784 against 0.756, interval $[-0.004, 0.058]$), and such differences are not claimed. Table~\ref{tab:suiteA} reports medians over framings, so its differences can exceed the canonical-row differences tested here. ECE noise floors at $n=500$ range from 0.002 (near-certain answers) to 0.049 (diffuse score questions), which is why raw ECE is never compared across questions.

\section{Full Suite A metrics}\label{app:full}
Table~\ref{tab:full} reports, for every Suite A question and checkpoint, accuracy (defined as in Table~\ref{tab:suiteA}) and the metrics of \S\ref{sec:metrics} not shown there: smooth calibration error, Brier score and clipped NLL on canonical-wording rows; coverage at 5\% risk; the argmax flip rate and mean pairwise Jensen--Shannon divergence across friendly framings; and, for score questions, the ranked probability score and quadratic weighted kappa.

\begin{table}[H]
\centering\small
\caption{Suite A ($n=500$ per generator). Accuracy: median and range over wordings. ECE/null: calibration error over its noise floor. $T$: refitted temperature ($>1$ over-confident; deg.: fit half all correct). p50: via API for \jev{} and the LLMs (per question), else local GPU. Baselines ignore the question wording; DistilBERT's 1.000 shows every label is learnable.}
\label{tab:suiteA}
\renewcommand{\arraystretch}{0.9}
\resizebox{\textwidth}{!}{%
\begin{tabular}{llcccccc}
\toprule
Question (primitive) & Model & Acc (median) & Range & ECE/null & $T$ & AURC & p50 ms \\
\midrule
\multirow{8}{*}{tickets.priority (score, 5)} & \jev{} 1.13.0 & {0.874} & [0.708, 0.946] & 2.2 & 0.66 & 0.023 & 233 \\
 & \laya{} english & 0.343 & [0.220, 0.372] & 3.3 & 5.04 & 0.660 & 59 \\
 & \laya{} typed-dec. & 0.383 & [0.347, 0.412] & 1.3 & 0.97 & 0.660 & 59 \\
 & \kev-0.8B & 0.282 & [0.236, 0.358] & 2.7 & 0.67 & 0.661 & 62 \\
 & \kev-4B & 0.446 & [0.382, 0.528] & 2.8 & 0.50 & 0.320 & 208 \\
 & \kev-9B & 0.565 & [0.476, 0.660] & 2.8 & 0.78 & 0.239 & 325 \\
  & Gemma 4 31B, constrained & 0.670 & [0.640, 0.710] & 25.4 & 5.13 & 0.137 & 542 \\
 & Qwen3.5-9B, constrained & 0.521 & [0.332, 0.584] & 1.9 & 1.15 & 0.303 & 631 \\
 & \textit{Baselines}: prior / regex / fine-tuned & \multicolumn{2}{c}{0.338 / 0.356 / 1.000} & -- & -- & -- & -- \\
\midrule
\multirow{8}{*}{phish.urgency (score, 4)} & \jev{} 1.13.0 & {1.000} & [1.000, 1.000] & 1.9 & deg. & 0.000 & 233 \\
 & \laya{} english & 0.462 & [0.462, 0.542] & 4.3 & 1.06 & 0.227 & 137 \\
 & \laya{} typed-dec. & 0.662 & [0.560, 0.682] & 4.0 & 0.27 & 0.093 & 137 \\
 & \kev-0.8B & 0.828 & [0.818, 0.910] & 8.3 & 0.32 & 0.021 & 131 \\
 & \kev-4B & {1.000} & [1.000, 1.000] & 4.6 & deg. & 0.000 & 489 \\
 & \kev-9B & {1.000} & [1.000, 1.000] & 4.6 & deg. & 0.000 & 677 \\
  & Gemma 4 31B, constrained & 1.000 & [1.000, 1.000] & 1.0 & deg. & 0.000 & 572 \\
 & Qwen3.5-9B, constrained & 1.000 & [1.000, 1.000] & 3.3 & deg. & 0.000 & 692 \\
 & \textit{Baselines}: prior / regex / fine-tuned & \multicolumn{2}{c}{0.296 / 1.000 / 1.000} & -- & -- & -- & -- \\
\midrule
\multirow{8}{*}{phish.is\_phishing (noul)} & \jev{} 1.13.0 & {0.984} & [0.984, 0.998] & 4.3 & 0.21 & 0.000 & 233 \\
 & \laya{} english & 0.550 & [0.486, 0.750] & 2.7 & 1.36 & 0.136 & 139 \\
 & \laya{} typed-dec. & 0.496 & [0.486, 0.640] & 7.0 & 2.15 & 0.203 & 140 \\
 & \kev-0.8B & 0.694 & [0.508, 0.766] & 4.1 & 0.75 & 0.099 & 139 \\
 & \kev-4B & 0.778 & [0.726, 0.790] & 4.6 & 0.31 & 0.046 & 513 \\
 & \kev-9B & 0.739 & [0.684, 0.756] & 3.4 & 0.71 & 0.063 & 777 \\
  & Gemma 4 31B, constrained & 0.969 & [0.960, 0.998] & 8.0 & 3.49 & 0.001 & 572 \\
 & Qwen3.5-9B, constrained & 0.996 & [0.924, 1.000] & 4.9 & deg. & 0.000 & 692 \\

 & \textit{Baselines}: prior / regex / fine-tuned & \multicolumn{2}{c}{0.514 / 0.514 / 1.000} & -- & -- & -- & -- \\
\midrule
\multirow{8}{*}{phish.attack\_class (choice, 5)} & \jev{} 1.13.0 & {1.000} & [0.984, 1.000] & 1.8 & deg. & 0.000 & 233 \\
 & \laya{} english & 0.755 & [0.588, 0.796] & 5.4 & 0.71 & 0.213 & 138 \\
 & \laya{} typed-dec. & 0.836 & [0.700, 0.884] & 10.2 & 0.23 & 0.016 & 139 \\
 & \kev-0.8B & 0.785 & [0.638, 0.896] & 4.6 & 0.61 & 0.074 & 137 \\
 & \kev-4B & 0.893 & [0.700, 0.944] & 3.9 & 0.36 & 0.009 & 508 \\
 & \kev-9B & 0.875 & [0.786, 0.950] & 2.3 & 0.62 & 0.024 & 767 \\
  & Gemma 4 31B, constrained & 0.936 & [0.902, 0.958] & 6.5 & 2.86 & 0.003 & 572 \\
 & Qwen3.5-9B, constrained & 0.987 & [0.944, 1.000] & 3.1 & 0.17 & 0.000 & 692 \\
 & \textit{Baselines}: prior / regex / fine-tuned & \multicolumn{2}{c}{0.486 / 0.898 / 1.000} & -- & -- & -- & -- \\
\bottomrule
\end{tabular}}
\end{table}

\begin{table}[h]
\centering\footnotesize
\caption{Full Suite A metrics ($n=500$ per generator). Acc: median accuracy over friendly wordings, as in Table~\ref{tab:suiteA}. Lower is better for smooth ECE, Brier, NLL, flip rate, JSD and RPS; higher is better for accuracy, coverage at 5\% risk and QWK. ``--'': not defined for the primitive.}
\label{tab:full}
\setlength{\tabcolsep}{3.5pt}
\begin{tabular}{llccccccccc}
\toprule
Question & Model & Acc & smooth ECE & Brier & NLL & cov@5\% & flip & JSD & RPS & QWK \\
\midrule
\texttt{tickets.queue} & \jev{} 1.13.0 & 1.000 & 0.015 & 0.004 & 0.016 & 1.00 & 0.029 & 0.026 & -- & -- \\
 & \laya{} english & 0.854 & 0.194 & 0.319 & 0.704 & 0.15 & 0.114 & 0.110 & -- & -- \\
 & \laya{} typed-dec. & 0.874 & 0.106 & 0.212 & 0.432 & 0.67 & 0.066 & 0.042 & -- & -- \\
 & \kev-0.8B & 1.000 & 0.128 & 0.038 & 0.146 & 1.00 & 0.045 & 0.079 & -- & -- \\
 & \kev-4B & 1.000 & 0.106 & 0.025 & 0.115 & 1.00 & 0.032 & 0.043 & -- & -- \\
 & \kev-9B & 1.000 & 0.100 & 0.024 & 0.113 & 1.00 & 0.009 & 0.016 & -- & -- \\
\midrule
\texttt{tickets.is\_angry} & \jev{} 1.13.0 & 0.997 & 0.055 & 0.009 & 0.057 & 1.00 & 0.003 & 0.011 & -- & -- \\
 & \laya{} english & 0.865 & 0.045 & 0.222 & 0.434 & 0.45 & 0.089 & 0.030 & -- & -- \\
 & \laya{} typed-dec. & 0.906 & 0.145 & 0.214 & 0.359 & 0.59 & 0.086 & 0.008 & -- & -- \\
 & \kev-0.8B & 0.797 & 0.282 & 0.188 & 0.355 & 1.00 & 0.203 & 0.024 & -- & -- \\
 & \kev-4B & 0.973 & 0.129 & 0.059 & 0.156 & 1.00 & 0.045 & 0.015 & -- & -- \\
 & \kev-9B & 0.994 & 0.082 & 0.030 & 0.098 & 1.00 & 0.012 & 0.013 & -- & -- \\
\midrule
\texttt{tickets.priority} & \jev{} 1.13.0 & 0.874 & 0.081 & 0.185 & 0.338 & 0.80 & 0.209 & 0.105 & 0.024 & 0.954 \\
 & \laya{} english & 0.343 & 0.130 & 0.832 & 2.003 & 0.00 & 0.400 & 0.072 & 0.169 & 0.342 \\
 & \laya{} typed-dec. & 0.383 & 0.049 & 0.721 & 1.374 & 0.00 & 0.288 & 0.010 & 0.129 & 0.116 \\
 & \kev-0.8B & 0.282 & 0.063 & 0.781 & 1.563 & 0.00 & 0.353 & 0.007 & 0.136 & 0.616 \\
 & \kev-4B & 0.446 & 0.114 & 0.617 & 1.153 & 0.08 & 0.292 & 0.018 & 0.087 & 0.773 \\
 & \kev-9B & 0.565 & 0.103 & 0.535 & 1.025 & 0.20 & 0.342 & 0.048 & 0.080 & 0.787 \\
\midrule
\texttt{phish.is\_phishing} & \jev{} 1.13.0 & 0.984 & 0.123 & 0.047 & 0.140 & 1.00 & 0.002 & 0.003 & -- & -- \\
 & \laya{} english & 0.550 & 0.072 & 0.359 & 0.536 & 0.29 & 0.227 & 0.110 & -- & -- \\
 & \laya{} typed-dec. & 0.496 & 0.200 & 0.507 & 0.696 & 0.28 & 0.060 & 0.015 & -- & -- \\
 & \kev-0.8B & 0.694 & 0.107 & 0.324 & 0.491 & 0.41 & 0.101 & 0.008 & -- & -- \\
 & \kev-4B & 0.778 & 0.160 & 0.304 & 0.481 & 0.66 & 0.057 & 0.007 & -- & -- \\
 & \kev-9B & 0.739 & 0.125 & 0.259 & 0.393 & 0.63 & 0.085 & 0.011 & -- & -- \\
\midrule
\texttt{phish.attack\_class} & \jev{} 1.13.0 & 1.000 & 0.014 & 0.002 & 0.014 & 1.00 & 0.006 & 0.011 & -- & -- \\
 & \laya{} english & 0.755 & 0.229 & 0.430 & 0.802 & 0.07 & 0.092 & 0.031 & -- & -- \\
 & \laya{} typed-dec. & 0.836 & 0.408 & 0.420 & 0.842 & 0.86 & 0.100 & 0.008 & -- & -- \\
 & \kev-0.8B & 0.785 & 0.180 & 0.301 & 0.642 & 0.40 & 0.115 & 0.028 & -- & -- \\
 & \kev-4B & 0.893 & 0.153 & 0.160 & 0.338 & 0.92 & 0.104 & 0.029 & -- & -- \\
 & \kev-9B & 0.875 & 0.085 & 0.176 & 0.341 & 0.78 & 0.070 & 0.028 & -- & -- \\
\midrule
\texttt{phish.urgency} & \jev{} 1.13.0 & 1.000 & 0.004 & 0.000 & 0.004 & 1.00 & 0.000 & 0.001 & 0.000 & 1.000 \\
 & \laya{} english & 0.462 & 0.201 & 0.582 & 1.104 & 0.33 & 0.045 & 0.010 & 0.158 & 0.319 \\
 & \laya{} typed-dec. & 0.662 & 0.154 & 0.419 & 0.800 & 0.54 & 0.107 & 0.002 & 0.102 & 0.743 \\
 & \kev-0.8B & 0.828 & 0.336 & 0.417 & 0.803 & 0.82 & 0.091 & 0.009 & 0.109 & 0.659 \\
 & \kev-4B & 1.000 & 0.130 & 0.042 & 0.148 & 1.00 & 0.000 & 0.006 & 0.016 & 1.000 \\
 & \kev-9B & 1.000 & 0.117 & 0.045 & 0.140 & 1.00 & 0.000 & 0.005 & 0.016 & 1.000 \\
\bottomrule
\end{tabular}
\end{table}

Table~\ref{tab:domaincal} gives the same calibration view for the four further generators.

\begin{table}[H]
\centering\footnotesize
\caption{Calibration on the four further generators (canonical wording, $n=300$): ECE divided by its noise floor, marked with the direction from the refitted temperature ($\uparrow$ over-confident, $\downarrow$ under-confident, $\approx$ calibrated, unmarked: degenerate fit), and AURC (lower is better). ``--'': every answer is correct at confidence 1, so ECE and its floor are both 0.}
\label{tab:domaincal}
\setlength{\tabcolsep}{3pt}
\resizebox{\textwidth}{!}{%
\begin{tabular}{l*{12}{c}}
\toprule
 & \multicolumn{2}{c}{\jev{} 1.13.0} & \multicolumn{2}{c}{\laya{} english} & \multicolumn{2}{c}{\laya{} typed-dec.} & \multicolumn{2}{c}{\kev-0.8B} & \multicolumn{2}{c}{\kev-4B} & \multicolumn{2}{c}{\kev-9B} \\
\cmidrule(lr){2-3}\cmidrule(lr){4-5}\cmidrule(lr){6-7}\cmidrule(lr){8-9}\cmidrule(lr){10-11}\cmidrule(lr){12-13}
Question & ECE/null & AURC & ECE/null & AURC & ECE/null & AURC & ECE/null & AURC & ECE/null & AURC & ECE/null & AURC \\
\midrule
logs.team (lookup) & -- & 0.000 & 2.5$\downarrow$ & 0.356 & 3.7$\downarrow$ & 0.359 & 4.3$\downarrow$ & 0.002 & 2.6$\downarrow$ & 0.000 & 2.0 & 0.000 \\
logs.severity (map) & 2.5 & 0.000 & 4.4$\uparrow$ & 0.628 & 1.7$\uparrow$ & 0.606 & 2.4$\downarrow$ & 0.197 & 4.4$\downarrow$ & 0.152 & 4.5$\downarrow$ & 0.052 \\
logs.page (2-field) & 3.4$\downarrow$ & 0.000 & 14.6$\uparrow$ & 0.804 & 6.5$\uparrow$ & 0.442 & 3.4$\downarrow$ & 0.063 & 3.7$\downarrow$ & 0.000 & 3.0$\approx$ & 0.016 \\
\midrule
policy.compliant & 3.7 & 0.000 & 1.9$\uparrow$ & 0.176 & 1.9$\downarrow$ & 0.194 & 2.1$\downarrow$ & 0.096 & 2.2$\downarrow$ & 0.028 & 1.4$\approx$ & 0.031 \\
policy.clause & 0.9 & 0.000 & 1.7$\uparrow$ & 0.636 & 2.0$\downarrow$ & 0.605 & 2.0$\downarrow$ & 0.175 & 2.0$\downarrow$ & 0.001 & 1.2$\downarrow$ & 0.006 \\
policy.severity & 2.7$\downarrow$ & 0.063 & 6.9$\uparrow$ & 0.586 & 3.4$\uparrow$ & 0.579 & 3.2$\downarrow$ & 0.259 & 1.9$\downarrow$ & 0.052 & 1.4$\downarrow$ & 0.048 \\
\midrule
guard.intent (4-way) & 2.8$\uparrow$ & 0.007 & 3.9$\uparrow$ & 0.490 & 4.4$\downarrow$ & 0.552 & 4.4$\downarrow$ & 0.221 & 2.5$\downarrow$ & 0.052 & 1.5$\downarrow$ & 0.031 \\
guard.block & 3.0 & 0.000 & 5.7$\uparrow$ & 0.236 & 4.3$\approx$ & 0.418 & 2.7$\downarrow$ & 0.161 & 2.6$\downarrow$ & 0.021 & 2.9 & 0.000 \\
guard.risk (score) & 1.9$\downarrow$ & 0.001 & 5.3$\uparrow$ & 0.543 & 6.2$\uparrow$ & 0.603 & 2.8$\downarrow$ & 0.332 & 2.9$\downarrow$ & 0.026 & 2.5$\downarrow$ & 0.071 \\
\midrule
multiling. queue & 1.3 & 0.000 & 4.5$\uparrow$ & 0.425 & 2.9$\approx$ & 0.355 & 2.9$\downarrow$ & 0.006 & 3.2 & 0.000 & 3.0 & 0.000 \\
multiling. anger & 3.3 & 0.000 & 4.5$\uparrow$ & 0.220 & 3.5$\uparrow$ & 0.216 & 4.5$\downarrow$ & 0.028 & 4.1$\downarrow$ & 0.000 & 3.5$\downarrow$ & 0.000 \\
multiling. priority & 3.4$\downarrow$ & 0.003 & 2.6$\uparrow$ & 0.632 & 1.9$\downarrow$ & 0.612 & 1.4$\uparrow$ & 0.702 & 2.9$\downarrow$ & 0.362 & 2.3$\downarrow$ & 0.369 \\
\bottomrule
\end{tabular}}
\end{table}

\section{Robustness, out-of-scope and prior shift}\label{app:robust}
Table~\ref{tab:robust} reports Suite D on the ticket set, with arm manifests paired item for item ($n=200$).

\begin{table}[H]
\centering\footnotesize
\caption{Suite D on the ticket set ($n=200$, paired arms). Largest accuracy change over the six surface perturbations and over the two distractor densities (which one and on which question in brackets). Out of scope: the 54 items whose true option is withheld; abstention when a ``none of the above'' option is offered, and without it the AUROC with which confidence separates these items from in-scope ones and the mean confidence on them versus in-scope items. Prior shift: mean $P(\text{yes})$ and accuracy on the anger question as its base rate is resampled to 0.2, 0.5 and 0.8.}
\label{tab:robust}
\setlength{\tabcolsep}{4pt}
\resizebox{\textwidth}{!}{%
\begin{tabular}{lccccccc}
\toprule
 & \multicolumn{2}{c}{Largest accuracy change} & \multicolumn{3}{c}{Out of scope} & \multicolumn{2}{c}{Prior shift (0.2 / 0.5 / 0.8)} \\
\cmidrule(lr){2-3}\cmidrule(lr){4-6}\cmidrule(lr){7-8}
Model & perturbation & distractors & abstain & AUROC & conf.\ out / in & mean $P(\text{yes})$ & accuracy \\
\midrule
\jev{} 1.13.0 & $-$0.060 (upper-case, priority) & $-$0.005 (0.25, priority) & 0.56 & 0.77 & 0.89 / 0.99 & 0.23 / 0.50 / 0.77 & 1.00 / 1.00 / 1.00 \\
\laya{} english & $-$0.305 (upper-case, queue) & $-$0.155 (0.5, queue) & 1.00 & 0.82 & 0.48 / 0.84 & 0.14 / 0.26 / 0.36 & 0.92 / 0.77 / 0.60 \\
\laya{} typed-dec. & $-$0.275 (upper-case, queue) & $-$0.135 (0.5, queue) & 0.76 & 0.93 & 0.43 / 0.78 & 0.26 / 0.36 / 0.45 & 0.94 / 0.85 / 0.73 \\
\kev-0.8B & $-$0.110 (upper-case, anger) & $-$0.030 (0.5, priority) & 0.96 & 0.97 & 0.55 / 0.89 & 0.39 / 0.52 / 0.65 & 0.97 / 0.99 / 0.99 \\
\kev-4B & $-$0.025 (5\% typos, anger) & $-$0.015 (0.25, priority) & 0.93 & 0.94 & 0.64 / 0.90 & 0.28 / 0.50 / 0.72 & 0.99 / 0.99 / 0.99 \\
\kev-9B & $-$0.055 (upper-case, priority) & $-$0.020 (0.5, priority) & 0.98 & 0.99 & 0.43 / 0.91 & 0.25 / 0.50 / 0.74 & 1.00 / 1.00 / 1.00 \\
\bottomrule
\end{tabular}}
\end{table}

\section{Ordinal ladders and yes/no consistency}\label{app:ordnoul}
Table~\ref{tab:ordnoul} reports Suites F and G for all six checkpoints.

\begin{table}[H]
\centering\footnotesize
\caption{Suites F and G. \emph{Severity ladders}: four fixed sequences of five customer complaints of increasing severity (Fig.~\ref{fig:app-ordinal}), each scored on 3-, 5- and 10-level rubrics; number of ladders (of four) on which the expected level never falls, exact accuracy of the most probable level, mean expected level at the mildest and the most severe step on the 5-level rubric (true levels 0 and 4), and Spearman correlation of the 20 complaints' scale positions between rubric sizes (range over the three pairs). \emph{Anger negation} ($n=500$ tickets): accuracy on the original question, and deviation $|P(\text{yes}\mid Q)+P(\text{yes}\mid\neg Q)-1|$ and accuracy for the explicit negation ``Is the customer calm, i.e.\ NOT expressing anger or hostility?'' and the paraphrase ``Is the tone of this message free of anger?''. \emph{Choice}: mean JSD and share of changed answers when the anger question is posed as a two-option choice. \emph{Threshold}: accuracy lost on phishing when the $P(\text{yes})$ threshold is fitted on anger rather than on phishing (500 items each). Phishing negations were run on the hosted model only: the explicit negation ``Is this email legitimate, i.e.\ NOT a phishing or social-engineering attack?'' deviates by 0.10 at accuracy 0.99, the paraphrase ``Is it safe to treat this message as genuine?'' by 0.23 at accuracy 0.76, against 0.99 on the original question.}
\label{tab:ordnoul}
\setlength{\tabcolsep}{3pt}
\resizebox{\textwidth}{!}{%
\begin{tabular}{l*{10}{c}}
\toprule
 & \multicolumn{4}{c}{Severity ladders (F)} & \multicolumn{3}{c}{Anger negation (G)} & \multicolumn{2}{c}{Choice (G)} & Threshold (G) \\
\cmidrule(lr){2-5}\cmidrule(lr){6-8}\cmidrule(lr){9-10}\cmidrule(lr){11-11}
Model & monotone 3/5/10 & exact 3/5/10 & level at 0 / 4 & Spearman & acc.\ original & dev.\ expl.\ / para. & acc.\ expl.\ / para. & JSD & changed & loss \\
\midrule
\jev{} 1.13.0 & 4 / 4 / 4 & 0.75 / 0.80 / 0.50 & 0.05 / 3.99 & 0.95--0.99 & 1.00 & 0.015 / 0.022 & 1.00 / 1.00 & 0.027 & 0.000 & 0.00 \\
\laya{} english & 3 / 1 / 1 & 0.55 / 0.45 / 0.20 & 1.22 / 2.96 & 0.81--0.91 & 0.84 & 0.604 / 0.756 & 0.19 / 0.12 & 0.012 & 0.050 & 0.06 \\
\laya{} typed-dec. & 2 / 2 / 2 & 0.65 / 0.45 / 0.35 & 1.33 / 3.18 & 0.89--0.96 & 0.90 & 0.382 / 0.478 & 0.28 / 0.10 & 0.014 & 0.084 & 0.23 \\
\kev-0.8B & 1 / 2 / 2 & 0.55 / 0.50 / 0.40 & 1.43 / 2.95 & 0.94--0.97 & 0.99 & 0.097 / 0.191 & 0.91 / 0.52 & 0.014 & 0.016 & 0.24 \\
\kev-4B & 2 / 3 / 3 & 0.90 / 0.75 / 0.55 & 0.52 / 3.51 & 0.96--0.98 & 0.99 & 0.036 / 0.031 & 0.99 / 0.98 & 0.001 & 0.002 & 0.11 \\
\kev-9B & 2 / 3 / 3 & 0.65 / 0.75 / 0.65 & 0.45 / 3.66 & 0.90--0.96 & 0.99 & 0.030 / 0.022 & 1.00 / 1.00 & 0.000 & 0.000 & 0.02 \\
\bottomrule
\end{tabular}}
\end{table}

\clearpage
\section{Hybrid routing}\label{app:hybrid}
Table~\ref{tab:hybrid} reports Suite H for both encoder checkpoints, escalating to the hosted model.

\begin{table}[H]
\centering\footnotesize
\caption{Suite H: the encoder answers every question, and answers whose confidence falls below the threshold $t$ are replaced by the hosted model's answers (one canonical-wording request per item; phishing: 500 items $\times$ 8 questions, tickets: 500 $\times$ 3, pooled). \emph{Alone}: the encoder's accuracy; \emph{AUROC}: how well its confidence separates its correct from its wrong answers (0.5 is chance). Per threshold: share of answers escalated, accuracy, accuracy when the same share is escalated at random, and share of requests with at least one escalated answer, which therefore need a hosted call. The hosted model alone scores 0.998 on phishing and 0.964 on tickets. By median billed tokens, a hosted request with one question costs 41\% (phishing) and 46\% (tickets) of a request with all of them.}
\label{tab:hybrid}
\begin{tabular}{llccccccc}
\toprule
Manifest & Encoder & alone & AUROC & $t$ & \shortstack{answers\\escalated} & accuracy & random & \shortstack{requests needing\\a hosted call} \\
\midrule
phishing & \laya{} english & 0.706 & 0.68 & 0.5 & 0.09 & 0.763 & 0.732 & 0.66 \\
 & & & & 0.7 & 0.34 & 0.858 & 0.804 & 1.00 \\
 & & & & 0.9 & 0.72 & 0.954 & 0.916 & 1.00 \\
\midrule
phishing & \laya{} typed-dec. & 0.762 & 0.74 & 0.5 & 0.15 & 0.823 & 0.798 & 0.87 \\
 & & & & 0.7 & 0.55 & 0.965 & 0.891 & 1.00 \\
 & & & & 0.9 & 0.96 & 0.998 & 0.988 & 1.00 \\
\midrule
tickets & \laya{} english & 0.692 & 0.74 & 0.5 & 0.22 & 0.779 & 0.753 & 0.63 \\
 & & & & 0.7 & 0.45 & 0.889 & 0.814 & 1.00 \\
 & & & & 0.9 & 0.61 & 0.919 & 0.857 & 1.00 \\
\midrule
tickets & \laya{} typed-dec. & 0.706 & 0.84 & 0.5 & 0.35 & 0.880 & 0.795 & 0.95 \\
 & & & & 0.7 & 0.59 & 0.945 & 0.857 & 1.00 \\
 & & & & 0.9 & 0.82 & 0.964 & 0.919 & 1.00 \\
\bottomrule
\end{tabular}
\end{table}

\section{Decision value}\label{app:dv}
Table~\ref{tab:dvall} extends Table~\ref{tab:dv} to every question with a cost matrix.

\begin{table}[H]
\centering\footnotesize
\caption{Suite I on all four questions with a cost matrix (one canonical-wording request per item, $n=500$.\ on tickets; cost per $10^4$ decisions). Value: cost(argmax) $-$ cost(Bayes) with a 95\% paired item-bootstrap interval (10{,}000 resamples); positive means acting on the probabilities pays. Cheapest constant action: the cost of always taking the single best action (for phishing, flagging every email). Cost matrices: phishing, a missed attack 20 and a false alarm 1; priority, 2 per level of under-prioritisation and 1 per level of over-prioritisation; anger, a missed angry customer 3 and a false alarm 0.5; queue, misrouting a billing or account ticket 3 and any other ticket 1. On 3{,}000 fresh phishing items the hosted model's value is $-$1{,}540, with interval [$-$2{,}487, $-$540] when items are resampled, but it is not significant when whole email templates are resampled: the loss comes from one template. Intervals here resample items; Table~\ref{tab:dv} resamples ticket templates, which widens them..}
\label{tab:dvall}
\begin{tabular}{llrrrc}
\toprule
Question & Model & argmax & Bayes & value & 95\% interval \\
\midrule
\texttt{phish.is\_phishing} & \multicolumn{5}{l}{cheapest constant action: 4{,}860} \\
 & \jev{} 1.13.0 & 3{,}200 & 4{,}820 & $-$1{,}620 & [$-$3{,}740, $+$820] \\
 & \laya{} english & 49{,}240 & 3{,}820 & $+$45{,}420 & [$+$37{,}800, $+$53{,}140] \\
 & \laya{} typed-dec. & 85{,}600 & 4{,}860 & $+$80{,}740 & [$+$71{,}780, $+$90{,}000] \\
 & \kev-0.8B & 58{,}000 & 4{,}860 & $+$53{,}140 & [$+$44{,}960, $+$61{,}380] \\
 & \kev-4B & 40{,}920 & 4{,}860 & $+$36{,}060 & [$+$28{,}940, $+$43{,}360] \\
 & \kev-9B & 48{,}800 & 4{,}860 & $+$43{,}940 & [$+$36{,}380, $+$51{,}940] \\
\midrule
\texttt{tickets.priority} & \multicolumn{5}{l}{cheapest constant action: 12{,}120} \\
 & \jev{} 1.13.0 & 1{,}080 & 2{,}300 & $-$1{,}220 & [$-$1{,}500, $-$940] \\
 & \laya{} english & 12{,}740 & 11{,}840 & $+$900 & [$+$280, $+$1{,}540] \\
 & \laya{} typed-dec. & 11{,}924 & 10{,}802 & $+$1{,}122 & [$+$441, $+$1{,}824] \\
 & \kev-0.8B & 13{,}580 & 9{,}860 & $+$3{,}720 & [$+$2{,}180, $+$5{,}240] \\
 & \kev-4B & 8{,}980 & 5{,}000 & $+$3{,}980 & [$+$3{,}000, $+$4{,}960] \\
 & \kev-9B & 4{,}780 & 7{,}960 & $-$3{,}180 & [$-$3{,}660, $-$2{,}720] \\
\midrule
\texttt{tickets.is\_angry} & \multicolumn{5}{l}{cheapest constant action: 3{,}560} \\
 & \jev{} 1.13.0 & 0 & 90 & $-$90 & [$-$150, $-$40] \\
 & \laya{} english & 4{,}680 & 2{,}050 & $+$2{,}630 & [$+$1{,}870, $+$3{,}460] \\
 & \laya{} typed-dec. & 3{,}066 & 2{,}004 & $+$1{,}062 & [$+$251, $+$1{,}904] \\
 & \kev-0.8B & 70 & 3{,}560 & $-$3{,}490 & [$-$3{,}690, $-$3{,}290] \\
 & \kev-4B & 100 & 1{,}320 & $-$1{,}220 & [$-$1{,}440, $-$980] \\
 & \kev-9B & 180 & 560 & $-$380 & [$-$610, $-$120] \\
\midrule
\texttt{tickets.queue} & \multicolumn{5}{l}{cheapest constant action: 11{,}800} \\
 & \jev{} 1.13.0 & 0 & 0 & 0 & [0, 0] \\
 & \laya{} english & 1{,}440 & 1{,}880 & $-$440 & [$-$660, $-$220] \\
 & \laya{} typed-dec. & 1{,}263 & 1{,}944 & $-$681 & [$-$902, $-$461] \\
 & \kev-0.8B & 0 & 280 & $-$280 & [$-$440, $-$140] \\
 & \kev-4B & 0 & 20 & $-$20 & [$-$60, 0] \\
 & \kev-9B & 0 & 120 & $-$120 & [$-$220, $-$40] \\
\bottomrule
\end{tabular}
\end{table}

\section{Additional figures}\label{app:figures}
\begin{figure}[tp]
\centering
\includegraphics{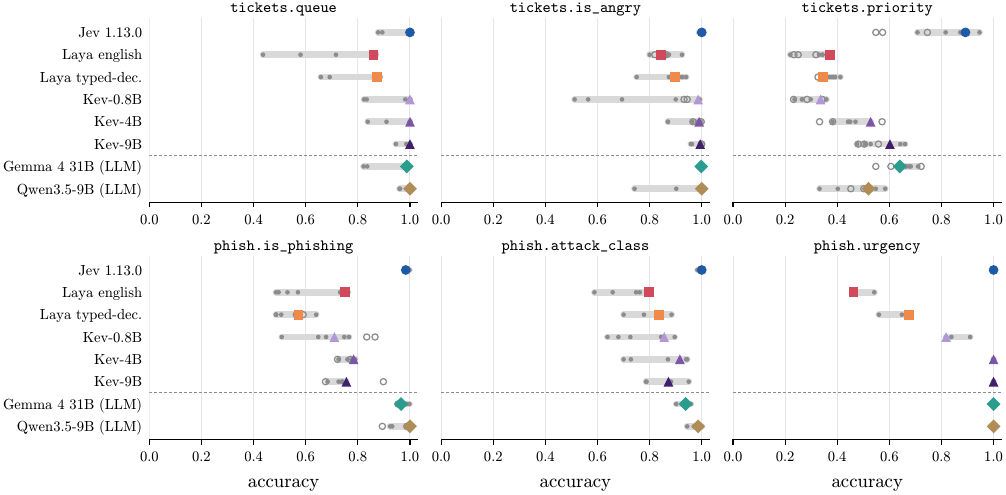}
\caption{Accuracy across framings for all six Suite A questions. Coloured marker: canonical wording; grey dots: paraphrases and criteria variants; open circles: adversarial wordings (ticket questions and phishing detection); grey bar: range over the friendly wordings. The saturated questions (queue, anger, attack class, urgency) show the criteria-variant floor at the left of each bar; priority shows the paraphrase spread. Rows below the dashed line: the two LLM baselines restricted to the answer options.}
\label{fig:app-framing}
\end{figure}

\begin{figure}[tp]
\centering
\includegraphics{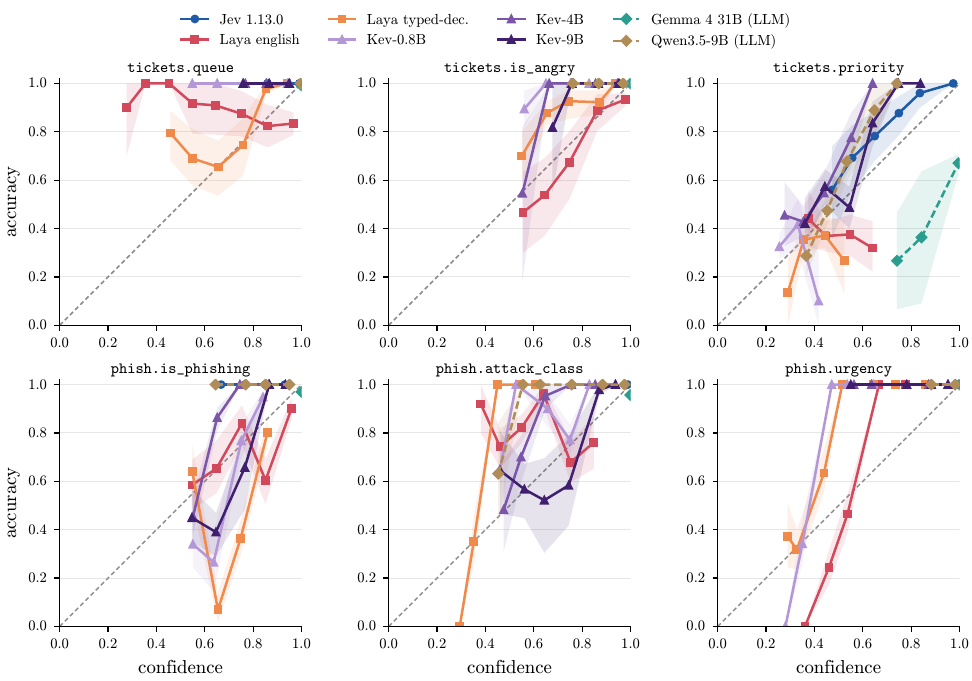}
\caption{Reliability diagrams (canonical wording, bootstrap bands per bin) for all six Suite A questions. Points above the diagonal indicate under-confidence. The hosted model and the larger decoders sit above the diagonal on every question; the encoder sits below it on the questions it gets wrong. Dashed: the two LLM baselines; Gemma 4 31B sits far below the diagonal on priority (over-confident).}
\label{fig:app-reliability}
\end{figure}

\begin{figure}[tp]
\centering
\includegraphics{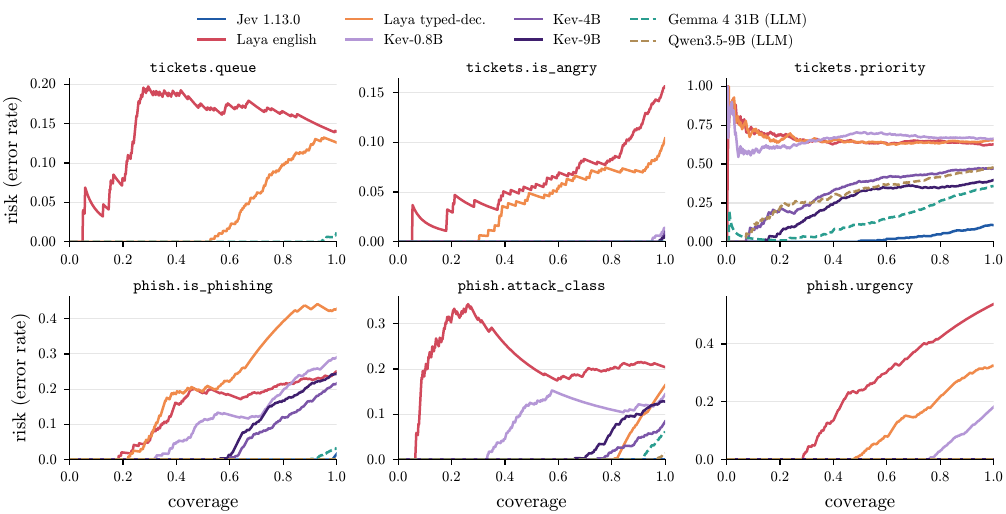}
\caption{Risk--coverage curves for all six Suite A questions: error rate among the items answered when answering only the most confident fraction (coverage). A curve that stays low until high coverage means confidence ranks errors well. The hosted model's curves are flat near zero on the saturated questions and rise only at full coverage on priority. Dashed: the two LLM baselines.}
\label{fig:app-rc}
\end{figure}

\begin{figure}[tp]
\centering
\includegraphics{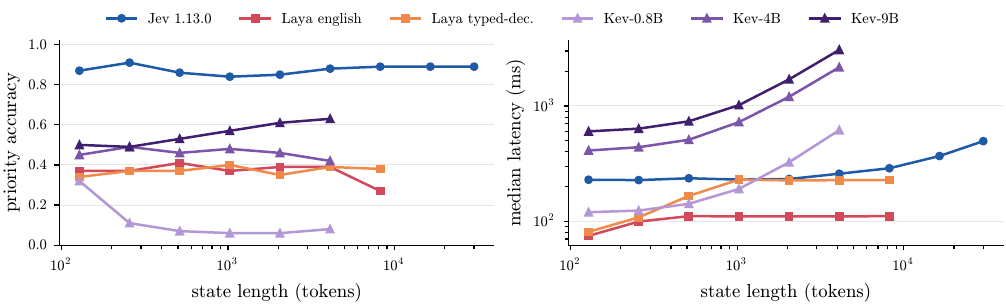}
\caption{Suite C, state length: priority accuracy (left) and median latency (right) as the state is padded from 128 to 30{,}000 tokens ($n=100$ per level). Levels at which a model refused most requests (the decoders' 8{,}192-token row limit) are omitted. The hosted model is flat in accuracy and sub-linear in latency; the encoder truncates silently and its English checkpoint drops at 8k tokens; the decoders' latency grows with length.}
\label{fig:app-length}
\end{figure}

\begin{figure}[tp]
\centering
\includegraphics{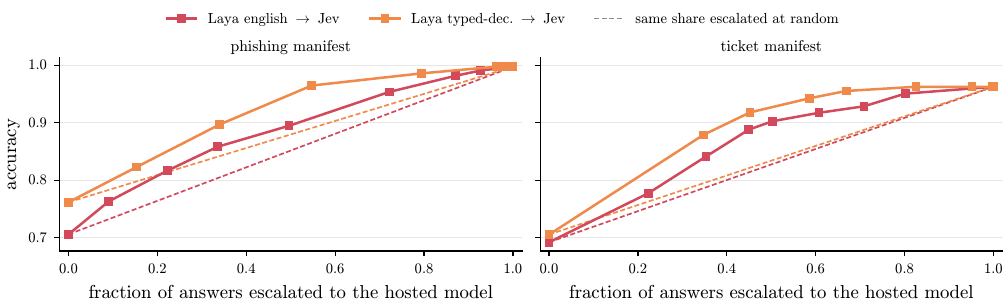}
\caption{Hybrid routing: accuracy (all questions of the manifest pooled) against the fraction of answers escalated from the encoder to the hosted model as the confidence threshold sweeps from 0.5 to 0.99. Dashed: escalating the same fraction of answers at random. Both encoder checkpoints lie above this line on both manifests, so their confidence picks out their errors better than chance (Table~\ref{tab:hybrid}).}
\label{fig:app-hybrid}
\end{figure}

\begin{figure}[tp]
\centering
\includegraphics{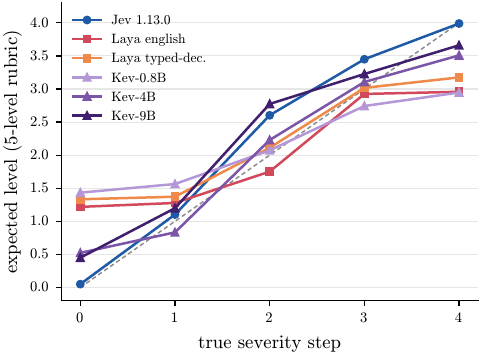}
\caption{Suite F, ordinal ladders on the 5-level rubric: mean expected level against the true severity step, averaged over four fixed ladders (dashed: the true level). Every averaged curve rises, but per ladder only the hosted model never falls (Table~\ref{tab:ordnoul}). The hosted model follows the true level; the 4B and 9B decoders use most of the scale; the 0.8B decoder and both encoders compress it, rating the mildest complaints too high and the most severe too low.}
\label{fig:app-ordinal}
\end{figure}

\begin{figure}[tp]
\centering
\includegraphics{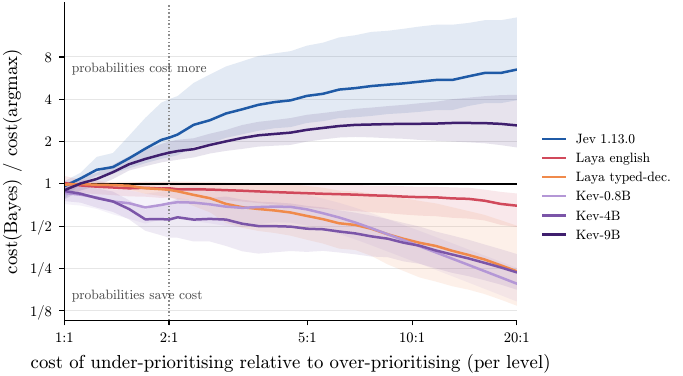}
\caption{Suite I, ticket priority as under-prioritising becomes more expensive than over-prioritising ($r$ per level against 1; dotted: the 2:1 matrix of Table~\ref{tab:dvall}). The curves show the cost of acting on the probabilities (Bayes) relative to acting on the top answer (argmax), on a log scale, for one canonical-wording request per item; above 1, acting on the probabilities costs more. Shaded: 95\% bands from resampling whole ticket templates (15), since items written from one template are not independent. For the hosted model and the 9B decoder, acting on the probabilities costs more from about 1.5:1 on, reaching 6.5 and 2.6 times the argmax cost at 20:1; for the other four checkpoints it saves cost, by more as the asymmetry grows (for the encoders, clearly only from about 5:1).}
\label{fig:app-costratio}
\end{figure}

\suppressfloats[t]
Figures~\ref{fig:app-framing}--\ref{fig:app-rc} repeat the Suite A plots of \S\ref{sec:results} for every question, on all six checkpoints and the two LLM baselines; Figures~\ref{fig:app-length}--\ref{fig:app-ordinal} plot the state-length sweep, the hybrid routing curves and the ordinal ladders that the main text only summarises, and Figure~\ref{fig:app-costratio} the decision value across cost ratios.


\section{Audit results}\label{app:audits}
State-only and options-only short circuits: on \texttt{tickets.queue}, options-only accuracy is at or below the majority prior (0.07--0.24) for every model; state-only accuracy with blank descriptions remains high (0.87 \jev, 0.47 \laya{} english, 0.91--0.92 \kev) because option keys such as \emph{billing} carry meaning, which the audit flags as expected. Positional bias over permutations: \jev{} chooses index 0 on 17.9\% of 6-option rows ($\chi^2$ $p<10^{-8}$), \laya{} on 22--23\% ($p<10^{-80}$), \kev{} on 17.7--18.0\%. Label-description leakage into the state is zero by construction for the ticket generator (states never name the queue). The label-noise control passed for every model. Label check: a frontier model (Claude Fable 5.1) given only the state and the question answered random samples of the evaluated manifests and matched every label on tickets (100 items, 300 answers), phishing (150 items, 1{,}200 answers), log triage, policy compliance and RAG relevance (80 items and 240 answers each). On guardrail intent it answered 110 of 150 items, the provider's content filter blocking 40 jailbreak, injection and harmful-request texts, and matched 320 of 330 labels. All ten disagreements fall on three templates whose stated rule leaves the label open: a role-play request for lock picking (risk 3 or 2, since lock picking is not on the rule's list of harms), a request to translate a quoted injection (block or answer) and an injected instruction to present a link as official (risk 2 or 3). These templates cover 20 of the 300 guardrail items; without them no entry of Table~\ref{tab:domains} changes by more than 0.06 (\kev-9B risk 0.903 $\to$ 0.961), and they are flagged for revision in the next generator release. The check runs before each generator release.
\end{document}